\documentclass[10pt, a4paper]{article}
\usepackage{lrec-coling2024} 

\usepackage{booktabs}
\usepackage{enumitem}
\usepackage{kotex}
\usepackage{multirow}
\usepackage{subcaption}
\usepackage{bigfoot}
\usepackage{dingbat}
\usepackage[breakable,skins]{tcolorbox}
\usepackage{amsmath}
\usepackage{lscape}
\usepackage{lipsum}
\usepackage[title,titletoc]{appendix}

\title{KoDialogBench: Evaluating Conversational Understanding of Language Models with Korean Dialogue Benchmark}

\name{Seongbo Jang\textsuperscript{\textdagger}\thanks{\textdagger These authors contributed equally to this work, which was done while the authors were at Scatter Lab.}, Seonghyeon Lee\textsuperscript{\textdagger}, Hwanjo Yu\textsuperscript{*}\thanks{*Corresponding author}} 

\address{Dept. of Computer Science and Engineering, POSTECH \\
         Pohang, South Korea \\
         \{\href{mailto:jang.sb@postech.ac.kr}{jang.sb}, \href{mailto:sh0416@postech.ac.kr}{sh0416}, \href{mailto:hwanjoyu@postech.ac.kr}{hwanjoyu}\}@postech.ac.kr\\}

\abstract{
As language models are often deployed as chatbot assistants, it becomes a virtue for models to engage in conversations in a user's first language.
While these models are trained on a wide range of languages, a comprehensive evaluation of their proficiency in low-resource languages such as Korean has been lacking.
In this work, we introduce KoDialogBench, a benchmark designed to assess language models' conversational capabilities in Korean.
To this end, we collect native Korean dialogues on daily topics from public sources, or translate dialogues from other languages.
We then structure these conversations into diverse test datasets, spanning from dialogue comprehension to response selection tasks.
Leveraging the proposed benchmark, we conduct extensive evaluations and analyses of various language models to measure a foundational understanding of Korean dialogues.
Experimental results indicate that there exists significant room for improvement in models' conversation skills.
Furthermore, our in-depth comparisons across different language models highlight the effectiveness of recent training techniques in enhancing conversational proficiency.
We anticipate that KoDialogBench will promote the progress towards conversation-aware Korean language models.
 \\ \newline 
 }

\begin{document}

\maketitleabstract

\section{Introduction}
\label{sec:introduction}
The recent advancement in large language models (LLMs) \citep{touvron2023llama,touvron2023llama2,chowdhery2022palm} has sparked an increased interest in evaluating their performance within the research community.
Several recent studies propose datasets to assess the abilities of language models in diverse ways \citep{cobbe2021training,bisk2020piqa,chen2021evaluating,zellers-etal-2019-hellaswag}.
Following this trend, the integration of these test sets into a unified benchmark has become crucial for a holistic evaluation of LLMs.
Notably, \citet{srivastava2023beyond,suzgun-etal-2023-challenging,eval-harness} curate benchmarks comprising diverse sets of real-world tasks through crowdsourcing, while \citet{hendrycks2021measuring} focus on evaluating general capabilities using regular exams.
These evaluations play a significant role in unveiling the functionalities of LLMs and transitioning LLMs to practical applications such as autonomous agents \citep{openai2023gpt4}.

Beyond these general functionalities, there also exists a rising interest in assessing LLMs for social interactions \citep{zhou2023far,wang2023emotional}.
Unlike conventional tasks that require logical knowledge, \citet{sap-etal-2019-social} emphasize the importance of commonsense reasoning for social interactions, and the subsequent work delves into the evaluation of these capabilities on LLMs \citep{sap-etal-2022-neural}.
Further, \citet{10.1145/3539618.3591877} construct a benchmark to analyze LLMs' understanding of social communications in the Chinese context.

However, there remains a notable gap between the evaluation protocols of LLMs for Korean language interactions and those for high-resource languages.
To the best of our knowledge, a comprehensive benchmark for assessing Korean conversational abilities of LLMs on daily topics has yet to be proposed.
Although \citet{NEURIPS2021_98dce83d} introduce a representative benchmark for Korean language understanding, it focuses on assessing the logical functionalities.
\citet{jang-etal-2022-kobest} construct another Korean benchmark designed by language experts, focusing on measuring the linguistic knowledge embedded in LLMs.
This lack of domain-specific evaluation methods potentially hinders the progression of Korean conversational LLMs.

In this work, we introduce KoDialogBench, a benchmark tailored to assess and compare the Korean conversational proficiency of LLMs.
To this end, we aggregate native Korean dialogues from public sources (e.g., AI Hub), or translate diverse open-domain dialogue corpora from other languages.
The collected dialogues are then framed into two primary tasks: dialogue comprehension and response selection.
We extensively leverage a variety of meta information provided by the original sources, facilitating a multifaceted analysis of conversational abilities.
Specifically, in dialogue comprehension, we probe various aspects to determine if a model is able to discern the underlying concepts within a dialogue.
For response selection, we evaluate a model's ability to distinguish appropriate next responses, categorizing dialogues by their metadata types.
Through these tasks, we aim to assess the depth of understanding and response accuracy of LLMs across diverse conversational scenarios.

Experimental results demonstrate that despite their extensive training on large-scale corpora, current LLMs fall short in matching human-level conversational abilities in Korean.
Although increasing the model size and incorporating well-curated Korean corpora during training improve performance, there still remains much room for LLMs to reach human-level understanding of open-domain dialogues.
Further analysis on heterogeneous dialogues discloses that most LLMs exhibit deficiencies in certain types of tasks, offering a precise diagnostic perspective for identifying areas of improvement.
Our benchmark not only furnishes a multifaceted viewpoints for assessing the conversational abilities of LLMs in Korean, but also paves the way for the development of adept conversational agents.\footnote{We make our code and data publicly available at \url{https://github.com/sb-jang/kodialogbench}.}

\section{Related Work}
\label{sec:relatedwork}
\paragraph{Dialogue Benchmarks}
With the advent of dialogue-based language models \citep{caldarini2022literature,10.1145/3166054.3166058,10.1007/s10462-022-10248-8}, a myriad of studies focus on evaluating these models in the context of open-domain dialogues.
Starting from a widely-used evaluation dataset in DSTC7 \citep{galley2019grounded} for response generation task, subsequent works have further enriched the field.
For instance, \citet{li-etal-2017-dailydialog} craft a dataset comprising multi-turn dialogues whose topics are related to daily life, and \citet{zhang-etal-2020-dialogpt} introduce a dataset derived from Reddit by transforming reply chains into dialogue structures.
Meanwhile, benchmarks designed to evaluate specific aspects of dialogues have also been proposed.
\citet{rashkin-etal-2019-towards} release a benchmark to assess the empathy exhibited by dialogue agents, and \citet{zhang-etal-2018-personalizing} scrutinize persona awareness through the lens of persona-guided dialogues.
\citet{shuster-etal-2020-dialogue} assemble diverse collections of open-domain dialogues, aiming to evaluate the capability of dialogue systems for engaging human-like conversations.
More recently, \citet{10.1145/3539618.3591877} present a benchmark intended to assess various social elements embedded within dialogues.
Nevertheless, a majority of these advancements have been confined to high-resource languages like English and Chinese, underlining a pertinent need for evaluation datasets for low-resource languages to assess conversational capabilities in a more diverse linguistic landscape.

\paragraph{Low-resource Language Benchmarks}
There exist several research works aiming at evaluating language models in the context of low-resource language understanding.
\citet{lai2023chatgpt,ahuja2023mega,bandarkar2023belebele,zhang2023m3exam,ryan-etal-2023-revisiting} employ a range of tasks across various languages to conduct multilingual evaluations.
While such multilingual assessments provide wide applicability for various languages, shifting a focus on a single specific language can yield a more sophisticated perspective and elevate the quality of the evaluation process.
For instance, \citet{uzunoglu-sahin-2023-benchmarking} focus on Turkish to provide a detailed evaluation of language model performance in that language.
Similarly, \citet{NEURIPS2022_890b206e} devise a benchmark exclusively for Polish to further analyze language models' understanding of the language.
Furthermore, \citet{kurihara-etal-2022-jglue} establish a benchmark centered on Japanese, employing original Japanese sentences.
Similar work has been conducted to measure the general understanding of Korean \citep{NEURIPS2021_98dce83d,jang-etal-2022-kobest}.
Nonetheless, a comprehensive evaluation of language models' proficiency in Korean conversational ability largely remains underexplored.

\section{Korean Dialogue Benchmark}
\label{sec:tasks}
\begin{table}[t]
\centering
\resizebox{\columnwidth}{!}{%
\begin{tabular}{@{}llrr@{}}
\toprule
\multicolumn{1}{c}{Task} & \multicolumn{1}{c}{Source} & \multicolumn{1}{c}{Class} & \multicolumn{1}{c}{Size} \\ \midrule
\multirow{3}{*}{Topic} & Korean SNS & 6 & 1200 \\
 & Korean Thematic Daily Dialogues & 19 & 1900 \\
 & SocialDial (Korean): Topic & 4 & 400 \\ \midrule
\multirow{3}{*}{Emotion} & Korean Emotional Dialogues & 6 & 1200 \\
 & DailyDialog (Korean): Emotion & 5 & 470 \\
 & Empathetic Dialogues (Korean): Emotion & 2 & 2000 \\ \midrule
\multirow{2}{*}{Relation} & SocialDial (Korean): Social Distance & 4 & 524 \\
 & SocialDial (Korean): Social Relation & 3 & 330 \\ \midrule
Location & SocialDial (Korean): Location & 4 & 376 \\ \midrule
\multirow{2}{*}{Dialog Act} & Korean Thematic Daily Dialogues & 4 & 520 \\
 & DailyDialog (Korean): Act & 4 & 1000 \\ \midrule
\multirow{3}{*}{Fact} & Korean Dialogue Summary & 4 & 1200 \\
 & PersonaChat (Korean): Persona & 4 & 1000 \\
 & Empathetic Dialogues (Korean): Situation & 4 & 2394 \\ \bottomrule
\end{tabular}%
}
\caption{Statistics for the dialogue comprehension task. Each task consists of one or more test sets, each with its own taxonomy.}
\label{tab:stats_dialogue_comprehension}
\end{table}

We construct a benchmark called KoDialogBench to evaluate the conversational capabilities of language models in Korean.
This benchmark consists of 21 test sets, encompassing diverse aspects of open-domain colloquial dialogues, categorized under two primary task suites: dialogue comprehension and response selection.
In this section, we outline the taxonomy of the benchmark and its construction process.
Throughout this process, we use held-out sources, specifically validation or test splits, to prevent benchmark contamination \citep{NEURIPS2020_1457c0d6,dodge-etal-2021-documenting,magar-schwartz-2022-data}.

\subsection{Dialogue Comprehension}
The dialogue comprehension task suite aims to quantify the ability of language models to identify diverse characteristics of conversations.
To this end, we exploit various meta information labeled to the dialogues, which spans from explicit information to implicit attributes inherent in the context.
The task suite encompasses six aspects and includes 14 test sets.
The taxonomy and statistics are presented in Table~\ref{tab:stats_dialogue_comprehension}.
Throughout our methodology, we exclude categories with less than 50 examples, applying stratified sampling to mitigate class imbalances.
Further details regarding these categories are elucidated in Appendix~\ref{appx:data_preprocessing}.

\subsubsection{Topic Classification}
Topic classification is widely used to assess whether language models can understand the main subject of a conversation \citep{guo2018topicbased}.
We leverage three publicly available corpora, each annotated with distinct topic categories.

We collect messenger chats from AI Hub website.\footnote{\url{https://bit.ly/3ZIUF3N}}
Out of the original categories, we select six classes that are clearly distinguishable and refer to this curated dataset as \textbf{Korean SNS}.
Similarly, we obtain Korean dialogues encompassing diverse topics.\footnote{\url{https://bit.ly/3ZNnqfG}}
We exclude the \textit{family} category from the original set due to its semantic overlap with other categories, resulting in the \textbf{Korean Thematic Daily Dialogues} dataset.
Finally, we acquire conversations from the SocialDial corpus~\cite{10.1145/3539618.3591877}.
We translate the native Chinese dialogues to Korean using the DeepL API.\footnote{For all translations in this study, we utilize the DeepL API. Link:~\url{https://bit.ly/3F2R1YM}}
We further exclude the \textit{life-trivial} category because of its overlapping semantics with other classes.
This modified dataset is denoted as \textbf{SocialDial (Korean): Topic}.

\subsubsection{Emotion Recognition}
Recognizing emotions is pivotal for engaging in social conversations~\cite{hsu-etal-2018-emotionlines,chatterjee-etal-2019-semeval}.
We harness three public sources to create classification datasets focused on this aspect.

From the AI Hub, we gather Korean human-bot counseling dialogues.\footnote{\url{https://bit.ly/3PFgOLK}}
This results in a dataset designed to gauge the counselee's emotions throughout the dialogue, named \textbf{Korean Emotional Dialogues}.
The major emotion labels based on six classes are used in this dataset.
Besides, we translate the DailyDialog~\cite{li-etal-2017-dailydialog} corpus, which has utterance-level emotion annotations.
We aim to identify the speaker's emotions in each utterance, resulting in the \textbf{DailyDialog (Korean): Emotion} dataset.
Lastly, we translate the Empathetic Dialogues~\cite{rashkin-etal-2019-towards} corpus, which contains dialogues grounded in emotional situations.
We binarize the original fine-grained emotions based on their polarity except two emotions, i.e., \textit{surprised} and \textit{sentimental}, whose polarities cannot be determined by their names.
This curated dataset, focusing on predicting a speaker's emotional polarity, is dubbed as \textbf{Empathetic Dialogues (Korean): Emotion}.

\subsubsection{Relation Classification}
Relation classification aims to discern the relations or the social distances between interlocutors~\cite{jia2021ddrel}.
The two test sets for this classification are derived from the translated SocialDial corpus, albeit with different class categories.

One aspect we focus on is the prediction of \textbf{social distance}, emphasizing the degree of closeness or acceptance between the interlocutors.
Additionally, we conceptualize the problem of determining \textbf{social relation}, which concentrates on the power distance between individuals.
We exclude dialogues that pertain to \textit{peer-peer} and \textit{elder-junior} categories due to formality inconsistency during translation~\cite{lee2023datadriven,lee-etal-2023-improving-formality}.

\subsubsection{Location Classification}
The goal of location classification is to determine where a dialogue takes place.
We also leverage the translated SocialDial corpus.
Out of the ten classes, the processed dataset exclude three classes that are difficult to infer from the given conversation alone.

\begin{table}[t]
\centering
\begin{tabular}{@{}lr@{}}
\toprule
\multicolumn{1}{c}{Source} & \multicolumn{1}{c}{Size} \\ \midrule
Korean SNS & 10295 \\
Korean Thematic Daily Dialogues & 10616 \\
Korean Emotional Dialogues & 17818 \\
PersonaChat (Korean) & 7801 \\
DailyDialog (Korean) & 6740 \\
Empathetic Dialogues (Korean) & 7941 \\
SocialDial (Korean) & 7237 \\ \bottomrule
\end{tabular}
\caption{Statistics for the response selection task data.}
\label{tab:stats_response_selection}
\end{table}

\subsubsection{Dialog Act Classification}
Understanding dialog acts is essential for creating socially adept conversational agents~\cite{stolcke-etal-2000-dialogue,shriberg-etal-2004-icsi}.
To this end, we leverage datasets from two distinct sources.

First, we utilize dialogues from the Korean Thematic Daily Dialogues, which come with coarse-grained dialog act classes such as \textit{directive}, \textit{assertive}, \textit{commissive}, and \textit{expressive}.
We devise descriptions for each class by integrating the verbalized text of fine-grained act labels and provide these descriptions in our prompts.
Additionally, we incorporate the translated DailyDialog to establish another test set.
Here, we follow the four act categories: \textit{inform}, \textit{question}, \textit{directive}, and \textit{commissive}.
We reference the explanations provided in the original work, translating them into Korean for use in our prompts.

\subsubsection{Fact Identification}
We devise three classification datasets that encompass varied facts within conversations, including dialogue summaries, personas, and situational contexts.

We collect conversations along with their summaries from AI Hub.\footnote{\url{https://bit.ly/3RL03RQ}}
Here, we construct queries presenting a four-option multiple choice format: one ground truth summary and three random distractors.
The objective is to correctly identify the summary that encapsulates the given dialogue.
This dataset is dubbed as \textbf{Korean Dialogue Summary}.

Additionally, we utilize a translated version of PersonaChat~\cite{zhang-etal-2018-personalizing} enriched with persona-grounded dialogues.
We construct four-option questions in which the ground truth is a persona sentence describing a speaker, and three distractors are persona sentences sampled from other examples.
This dataset is named \textbf{PersonaChat (Korean): Persona}.

Lastly, for discerning situational contexts, we employ the translated Empathetic Dialogues corpus.
The dataset contains descriptions detailing the situations in which the dialogues occur.
We formulate the problem that involves presenting a ground truth situation alongside three distractors sampled from other dialogues.

\subsection{Response Selection}
We construct seven response selection datasets from our collection of Korean conversations through the following processes.
For Korean SNS, we sample the same number of conversations based on attributes like the number of participants, the gender composition of dialogues, and dialogue topics.
This not only reduces the computational costs for evaluation \citep{maynez-etal-2023-benchmarking}, but also ensures fair representation of various dialogues.
For corpora with more than 10k dialogues, namely Korean SNS and Korean Thematic Daily Dialogues, we sample informative utterances based on character count and the number of unique Korean characters.
Using these sampled utterances, we create examples that require a model to predict these utterances.
For Korean Emotional Dialogues, we craft instances for bot responses to measure the capabilities of models to empathize with humans.
For the remaining corpora, every utterance is transformed into a response selection example.

To curate five-option multiple choice questions, we randomly sample four responses from the same corpus to act as negatives.
The statistics of the datasets are presented in Table~\ref{tab:stats_response_selection}.

\section{Experiments}
\label{sec:experiments}
\subsection{Experimental Setup}

\begin{table*}[t]
\centering
\begin{tabular}{@{}llcccc@{}}
\toprule
\multicolumn{1}{c}{Model} & \multicolumn{1}{c}{Base} & Korean & Chinese & Code & Instruction \\ \midrule
XGLM \citep{lin-etal-2022-shot} &  &  &  &  & \\
LLaMA \citep{touvron2023llama} &  &  &  &  & \\
WizardLM \citep{xu2023wizardlm} & LLaMA &  &  &  & \checkmark \\
LLaMA-2 \citep{touvron2023llama2} &  &  &  &  & \\
LLaMA-2-Chat \citep{touvron2023llama2} & LLaMA-2 &  &  &  & \checkmark \\
Falcon \citep{penedo2023refinedweb} &  &  &  &  & \\
Falcon-Inst \citep{penedo2023refinedweb} & Falcon &  &  &  & \checkmark \\
Mistral \citep{jiang2023mistral} &  &  &  &  & \\
Mistral-Inst \citep{jiang2023mistral} & Mistral &  &  &  & \checkmark  \\
CodeLLaMA \citep{roziere2023code} & LLaMA-2 &  &  & \checkmark & \\
CodeLLaMA-Inst \citep{roziere2023code} & CodeLLaMA &  &  & \checkmark & \checkmark \\
Qwen \citep{bai2023qwen} &  &  & \checkmark  &  & \\
Qwen-Chat \citep{bai2023qwen} & Qwen &  & \checkmark  &  & \checkmark \\
Polyglot-Ko \citep{ko2023technical} &  & \checkmark &  &  & \\
KoAlpaca & PolyGlot-Ko & \checkmark &  &  & \checkmark \\
KORani-v1 & PolyGlot-Ko & \checkmark &  &  & \checkmark \\
KORani-v2 & LLaMA & \checkmark &  &  & \checkmark \\
KORani-v3 & LLaMA & \checkmark &  &  & \checkmark \\ \bottomrule
\end{tabular}
\caption{Details of language models used in experiments. Korean and Chinese indicate whether the majority of pretraining text is Korean or Chinese, respectively. Code and Instruction denote whether the model is additionally trained using code corpora or instruction datasets, respectively.}
\label{tab:experiment-model-detail}
\end{table*}

\subsubsection{Language Models}
In Table~\ref{tab:experiment-model-detail}, we organize recently published language models based on four criteria that can influence Korean dialogue tasks.
Detailed attributes related to their handling Korean are as follows:
\begin{itemize}[leftmargin=10pt]
    \setlength{\itemsep}{0pt}
    \item LLaMA-2 is pretrained using Korean texts; however, they comprise less than 0.1\% of the entire corpus.
    \item Polyglot-Ko is pretrained on a Korean corpus collected from a variety of sources including dialogue data such as ClovaCall \citep{ha2020clovacall}.
    \item KoAlpaca utilizes the Alpaca dataset \citep{alpaca}, which is translated into Korean for instruction tuning.
    \item KORani adopts a similar approach as KoAlpaca but uses the Vicuna dataset \citep{zheng2023judging}.
\end{itemize}

\subsubsection{Evaluation Protocols}
We adopt the multiple-choice format, which is prevalent for evaluating language models \citep{hendrycks2021measuring,eval-harness}.
In this approach, a language model calculates the log-likelihood of generating each option given a prompt and makes a selection accordingly.
To ensure language models effectively focus on the target tasks, we employ several prompting strategies.
Illustrative examples of these prompting methods are presented in the Appendix~\ref{appx:prompts}.

\paragraph{Direct Prompting}
Direct prompting requires a model to generate answers directly for the presented problem.
The prompt consists of a dialogue and a subsequent question regarding the characteristics of the dialogue.
We take the log probabilities of each verbalized class name as its scores.
This method is applied for topic, emotion, relation, and location classification, as the verbalized class names aptly convey their meanings.

\paragraph{Direct Prompting with Class Descriptions}
For tasks where class names alone are not sufficiently descriptive, we supplement the prompt with short explanations about each class.
This method is employed for dialog act classification, as the class names are often too abstract for models to solve the task accurately.

\paragraph{Option Prompting}
Option prompting displays options alongside their corresponding numbers within the prompt.
A model is then asked to generate the number corresponding to the correct class.
We adopt this method for fact identification since the candidates differ for each example.

\paragraph{Response Selection Prompting}
Response selection can be structured as a text completion task, which is a format well-suited for language models.
The prompt is organized as a sequence of utterances, ending with the speaker identifier for the subsequent utterance.
Given this prompt, it is natural for a model to generate the next utterance, thereby completing the dialogue structure.
It can be interpreted that the model is inclined to generate responses with the highest score compared to the other options.

\begin{table*}[t]
\centering
\begin{tabular}{@{}lccccccc@{}}
\toprule
\multicolumn{1}{c}{Model} & Topic & Location & Relation & Emotion & Dialog Act & Fact & Average \\ \midrule
Random & 15.6 & 25.0 & 29.2 & 28.9 & 25.0 & 25.0 & 24.8 \\ \midrule
XGLM 564M & 30.8 & 52.1 & 35.8 & 37.1 & 24.7 & 25.2 & 34.3 \\
XGLM 1.7B & 30.2 & 48.1 & 33.4 & 39.5 & 25.0 & 25.2 & 33.6 \\
XGLM 2.9B & 37.2 & 45.7 & 41.1 & 44.3 & 25.0 & 25.2 & 36.4 \\
XGLM 4.5B & 32.3 & 61.4 & 36.0 & 43.5 & 25.8 & 25.2 & 37.4 \\
XGLM 7.5B & 38.9 & 69.4 & 42.3 & 50.5 & 24.9 & 25.0 & 41.8 \\
LLaMA 7B & 26.2 & 35.9 & 43.9 & 46.1 & 23.7 & 26.1 & 33.7 \\
LLaMA 13B & 29.3 & 42.3 & 37.5 & 37.7 & 23.6 & 34.2 & 34.1 \\
WizardLM 7B & 15.6 & 25.0 & 29.2 & 28.9 & 25.0 & 24.3 & 24.7 \\
WizardLM 13B & 28.6 & 42.6 & 36.4 & 36.1 & 24.6 & 34.5 & 33.8 \\
LLaMA-2 7B & 37.5 & 75.8 & 56.6 & 46.4 & 24.7 & 32.1 & 45.5 \\
LLaMA-2 13B & 36.0 & 78.2 & 53.3 & 54.6 & 24.8 & 38.0 & 47.5 \\
LLaMA-2-Chat 7B & 31.0 & 67.3 & 44.3 & 43.3 & 27.1 & 32.7 & 41.0 \\
LLaMA-2-Chat 13B & 37.0 & 74.7 & 46.2 & 51.3 & 23.6 & 41.5 & 45.7 \\
Falcon 7B & 19.7 & 29.5 & 35.1 & 32.9 & 25.0 & 25.2 & 27.9 \\
Falcon-Inst 7B & 22.5 & 26.6 & 37.2 & 36.3 & 24.6 & 25.2 & 28.7 \\
Mistral 7B & 34.1 & 76.9 & 46.7 & 58.8 & 25.0 & 68.3 & 51.6 \\
Mistral-Inst 7B & 27.9 & 39.9 & 52.9 & 42.8 & 26.9 & 53.0 & 40.6 \\
CodeLLaMA 7B & 30.7 & 52.7 & 39.8 & 46.7 & 23.8 & 41.8 & 39.2 \\
CodeLLaMA 13B & 33.8 & 63.3 & 54.0 & 63.1 & 25.5 & 31.8 & 45.3 \\
CodeLLaMA-Inst 7B & 32.1 & 55.3 & 41.5 & 52.2 & 26.8 & 47.2 & 42.5 \\
CodeLLaMA-Inst 13B & 32.9 & 63.3 & 57.5 & 63.9 & 26.0 & 49.4 & 48.8 \\
Qwen 7B & 38.8 & 64.1 & 29.4 & 49.4 & 25.0 & 45.1 & 42.0 \\
Qwen 14B & 45.3 & 73.7 & 41.9 & 59.8 & 27.7 & 77.2 & 54.3 \\
Qwen-Chat 7B & 36.2 & 48.4 & 38.5 & 37.2 & 25.4 & 58.0 & 40.6 \\
Qwen-Chat 14B & 41.8 & 68.6 & 41.5 & 54.8 & 28.7 & 82.9 & 53.0 \\
Polyglot-Ko 1.3B & 31.8 & 61.7 & 39.1 & 44.8 & 24.8 & 25.3 & 37.9 \\
Polyglot-Ko 3.8B & 36.2 & 58.8 & 45.4 & 48.1 & 25.2 & 25.3 & 39.8 \\
Polyglot-Ko 5.8B & 29.7 & 59.3 & 40.0 & 46.5 & 26.3 & 25.3 & 37.8 \\
Polyglot-Ko 12.8B & 36.7 & 61.7 & 47.2 & 53.9 & 24.9 & 24.3 & 41.5 \\
KoAlpaca 5.8B & 33.1 & 47.1 & 31.6 & 40.3 & 23.3 & 24.2 & 33.3 \\
KoAlpaca 12.8B & 42.1 & 70.5 & 44.2 & 60.1 & 24.3 & 23.7 & 44.1 \\
KORani-v1 13B & 33.4 & 73.1 & 45.8 & 52.0 & 24.0 & 24.3 & 42.1 \\
KORani-v2 13B & 30.5 & 68.1 & 45.7 & 39.0 & 25.3 & 34.6 & 40.5 \\
KORani-v3 13B & 34.7 & 69.7 & 41.5 & 48.3 & 26.7 & 36.1 & 42.8 \\ \midrule
Human & 83.6 & 86.0 & 73.3 & 67.1 & 54.7 & 93.3 & 76.3 \\ \bottomrule
\end{tabular}
\caption{Results for dialogue comprehension. Scores for each task represent the average across test sets within a task. Detailed results for individual tasks are presented in Appendix~\ref{appx:detailed_results}.}
\label{tab:result-comprehension}
\end{table*}

\begin{table*}[t]
\centering
\begin{tabular}{@{}lcccccccc@{}}
\toprule
\multicolumn{1}{c}{Model} & K-SNS & K-TDD & K-ED & PC (K) & DD (K) & ED (K) & SD (K) & Average \\ \midrule
XGLM 564M & 22.3 & 24.0 & 38.0 & 30.1 & 30.5 & 25.9 & 32.9 & 29.1 \\
XGLM 1.7B & 23.4 & 27.1 & 42.1 & 33.0 & 33.6 & 28.1 & 35.4 & 31.8 \\
XGLM 2.9B & 24.0 & 29.1 & 45.6 & 34.9 & 34.9 & 29.4 & 38.1 & 33.7 \\
XGLM 4.5B & 22.9 & 26.9 & 42.6 & 32.8 & 33.7 & 28.3 & 35.5 & 31.8 \\
XGLM 7.5B & 25.3 & 31.1 & 47.5 & 36.4 & 36.4 & 30.9 & 39.5 & 35.3 \\
LLaMA 7B & 20.4 & 22.2 & 38.1 & 31.0 & 28.9 & 24.5 & 30.3 & 29.2 \\
LLaMA 13B & 21.1 & 23.4 & 39.5 & 33.4 & 30.4 & 25.7 & 32.1 & 29.4 \\
WizardLM 7B & 15.4 & 13.0 & 21.0 & 21.0 & 22.2 & 20.7 & 23.0 & 19.5 \\
WizardLM 13B & 22.1 & 24.7 & 41.5 & 35.0 & 31.5 & 27.5 & 33.9 & 30.9 \\
LLaMA-2 7B & 22.1 & 25.5 & 41.7 & 35.1 & 33.0 & 27.7 & 35.6 & 31.5 \\
LLaMA-2 13B & 23.4 & 28.1 & 44.1 & 36.9 & 34.7 & 28.9 & 37.2 & 33.3 \\
LLaMA-2-Chat 7B & 22.6 & 25.3 & 41.3 & 35.3 & 33.5 & 28.4 & 36.3 & 31.8 \\
LLaMA-2-Chat 13B & 23.6 & 27.4 & 42.1 & 35.6 & 34.3 & 28.1 & 36.4 & 32.5 \\
Falcon 7B & 20.0 & 20.4 & 36.4 & 29.2 & 26.2 & 23.6 & 28.0 & 26.3 \\
Falcon-Inst 7B & 18.9 & 19.1 & 35.1 & 26.7 & 25.2 & 23.3 & 27.3 & 25.1 \\
Mistral 7B & 24.4 & 28.9 & 46.3 & 37.6 & 35.2 & 29.7 & 38.2 & 34.3 \\
Mistral-Inst 7B & 22.5 & 25.9 & 41.9 & 35.8 & 33.1 & 27.7 & 34.9 & 31.7 \\
CodeLLaMA 7B & 23.8 & 27.4 & 42.1 & 35.6 & 34.3 & 28.1 & 36.4 & 32.5 \\
CodeLLaMA 13B & 24.8 & 28.5 & 43.8 & 37.0 & 34.9 & 29.6 & 38.2 & 33.8 \\
CodeLLaMA-Inst 7B & 23.8 & 27.5 & 41.9 & 36.3 & 34.2 & 28.7 & 37.1 & 32.8 \\
CodeLLaMA-Inst 13B & 24.8 & 29.4 & 44.2 & 38.3 & 35.6 & 30.2 & 39.2 & 34.5 \\
Qwen 7B & 22.3 & 25.2 & 41.4 & 35.2 & 33.7 & 27.8 & 36.3 & 31.7 \\
Qwen 14B & 25.4 & 31.0 & 49.4 & 39.9 & 37.7 & 32.2 & 41.9 & 36.8 \\
Qwen-Chat 7B & 22.8 & 26.2 & 43.1 & 35.9 & 33.9 & 28.4 & 36.2 & 32.3 \\
Qwen-Chat 14B & 25.8 & 31.9 & 50.6 & 40.6 & 38.9 & 33.2 & 43.3 & 37.8 \\
Polyglot-Ko 1.3B & 28.9 & 33.7 & 46.3 & 33.9 & 35.5 & 30.7 & 39.5 & 35.5 \\
Polyglot-Ko 3.8B & 31.4 & 37.3 & 50.0 & 35.8 & 37.7 & 31.8 & 41.6 & 37.9 \\
Polyglot-Ko 5.8B & 31.0 & 38.0 & 50.8 & 35.3 & 37.7 & 32.0 & 41.9 & 38.1 \\
Polyglot-Ko 12.8B & 33.6 & 41.0 & \textbf{54.2} & 36.4 & 39.1 & 32.6 & 43.2 & 40.0 \\
KoAlpaca 5.8B & 25.7 & 29.1 & 39.9 & 31.5 & 33.5 & 29.3 & 38.0 & 32.4 \\
KoAlpaca 12.8B & 33.3 & 38.8 & 49.3 & 36.0 & 37.8 & 32.6 & 42.5 & 38.6 \\
KORani-v1 13B & \textbf{35.8} & \textbf{43.0} & 53.2 & \textbf{41.7} & \textbf{41.0} & \textbf{35.6} & \textbf{45.6} & \textbf{42.3} \\
KORani-v2 13B & 22.1 & 25.4 & 41.0 & 39.4 & 37.1 & 32.3 & 39.9 & 33.9 \\
KORani-v3 13B & 22.1 & 25.2 & 40.9 & 39.6 & 37.3 & 32.2 & 39.6 & 33.8 \\ \midrule
Human & 84.0 & 98.0 & 98.7 & 86.0 & 90.0 & 87.3 & 88.7 & 90.4 \\ \bottomrule
\end{tabular}
\caption{Results for response selection. K-SNS: Korean SNS and Dialogue Summary, K-TDD: Korean Thematic Daily Dialogues, K-ED: Korean Emotional Dialogues, PC (K): PersonaChat (Korean), DD (K): DailyDialog (Korean), ED (K): Empathetic Dialogues (Korean), and SD (K): SocialDial (Korean).}
\label{tab:result-response-selection}
\end{table*}

\subsection{Results}
We report the accuracy results for the two task suites in Table~\ref{tab:result-comprehension} and Table~\ref{tab:result-response-selection}.
Our analysis delves into the empirical findings related to the Korean conversational capabilities of the models.

\paragraph{Model Scaling}
We observe model scaling trends across all model groups in both task suites.
Generally, larger models outperform their smaller counterparts, with the exception of Polyglot-Ko 5.8B and XGLM 4.5B.
This suggests that increasing the model size may be an effective strategy to enhance Korean conversational capabilities, but solely scaling the model does not guarantee significant improvement.

\paragraph{Cross-lingual Transferability of Instruction Tuning}
We find that instruction tuning using datasets in languages other than Korean does not improve Korean conversational capabilities.
Specifically, WizardLM, LLaMA-2-Chat, Qwen-Chat, and Mistral-Inst show lower average accuracy scores in dialogue comprehension and show little to no improvement in response selection compared to their base models: LLaMA, LLaMA-2, Qwen, and Mistral, respectively.
On the other hand, instruction tuning with datasets in Korean such as KoAlpaca and KORani, improve scores despite potential translation errors.
This indicates that the general improvement coming from instruction tuning, as discussed in several research \citep{wei2022finetuned,chung2022scaling}, does not exhibit cross-lingual transferability.

\paragraph{Effects of Instruction Tuning Datasets}
KoAlpaca models exhibit proficiency in dialogue comprehension tasks, whereas KORani-v1 excels in response selection tasks, although both models are fine-tuned on the same base model, Polyglot-Ko.
We attribute this distinction to the semantic difference of training examples in their respective instruction tuning datasets.
The Alpaca dataset primarily consists of question-answering dialogues, which naturally aligns with dialogue comprehension tasks that focus on understanding dialogue characteristics.
In contrast, the Vicuna dataset encompasses more realistic conversations, potentially enabling the model to preserve the capabilities of responding to diverse conversations.
This suggests that while instruction-tuning is thought to enhance the generalization across various tasks, the format of the instructions also influences the performance of the target task.

\paragraph{Training with Code Data}
We observe that code pretraining increases the ability to identify facts within Korean conversations.
As evidenced in Table~\ref{tab:result-comprehension}, CodeLLaMA-Inst outperforms LLaMA-2-Chat in fact identification.
This aligns with the observations of \citet{madaan-etal-2022-language}, indicating that code training boosts reasoning capabilities.
It also brings improvements to response selection performances with the same models.

\paragraph{Pretraining with Large Proportion of Korean Corpus}
Our experimental results demonstrate that language models primarily pretrained on a large-scale Korean corpus, specifically Polyglot-Ko, show better conversational proficiency compared to other models.
The most competitive model is Qwen-Chat 14B, but it exhibits lower accuracy scores on tasks derived from native Korean conversations, namely K-SNS, K-TDD, and K-ED.
Considering that the Korean text proportion in the pretraining dataset of LLaMA-2 is less than 0.1\%, we speculate that such a proportion is insufficient to capture the intrinsic nuances and cultural context of Korean dialogues.
Meanwhile, in dialogue comprehension tasks, some multilingual models like LLaMA-2, Mistral, and Qwen outperform Polyglot-Ko.
This indicates that while these models encode the basic understanding of Korean conversations, their adeptness in generating appropriate responses remains limited.

\paragraph{Fine-tuning with Korean Data}
It is worth noting the potential of fine-tuning on Korean data for dialogue tasks.
Both KORani-v2 and KORani-v3 consistently outperform their base model LLaMA on both task suites.
This implies that additional training on Korean texts elicits a model's capability in Korean conversation, even though Korean is not primarily utilized during pretraining.
However, significant improvements are seen in the response selection tasks composed of translated conversations, namely PC (K), DD (K), ED (K), and SD (K).
Therefore, more sophisticated fine-tuning in Korean is essential to effectively harness the Korean conversational ability embedded in language models.

\paragraph{Human Performance}
We find that current state-of-the-art language models still lag behind human performance across various tasks.
To measure human performance, we employ three native speakers and have them solve 50 randomly sampled problems per each task.\footnote{Fleiss' $\kappa = 0.793$ for all tasks, which indicates substantial agreements. The results for each task are detailed in Appendix~\ref{appx:detailed_results}.}
We present the average accuracy scores of these three participants.
As a result, we observe that there exists a large performance gap between the models and human participants in both task suites.
This signifies that there remains room for further improving the models' proficiency in Korean conversations.

\begin{table}[t]
\centering
\begin{tabular}{@{}lcc@{}}
\toprule
\multicolumn{1}{c}{Model} & Bilateral & Multilateral \\ \midrule
XGLM 7.5B & 26.8 & 23.5 \\
LLaMA 13B & 22.7 & 19.5 \\
LLaMA-2 13B & 25.2 & 21.5 \\
Falcon 7B & 21.5 & 18.4 \\
Mistral 7B & 26.2 & 22.4 \\
CodeLLaMA 13B & 26.3 & 23.1 \\
Qwen 14B & 27.7 & 22.9 \\
Polyglot-Ko 12.8B & 35.7 & 31.2 \\
KORani-v1 13B & 38.4 & 32.9 \\ \midrule
Human & 84.0 & 82.6 \\ \bottomrule
\end{tabular}
\caption{Response selection accuracy for bilateral and multilateral dialogues. Dataset: K-SNS.}
\label{tab:response-selection-number-of-participant}
\end{table}

\paragraph{Analysis on the Number of Speakers}
We further investigate the effects of the number of speakers on model performance.
We evaluate response selection accuracy for two dialogue types: bilateral and multilateral, using the K-SNS dataset.
The results are reported in Table~\ref{tab:response-selection-number-of-participant}.
All models show higher accuracy with bilateral dialogues as opposed to multilateral dialogues.
In contrast, human performance remains similar across both dialogue types.
This implies that language models struggle to accurately trace the interlocutors' information as the number of speakers increases, which is consistent with the finding in prior work \citep{sap-etal-2022-neural}.
This trend is also observed across all models, highlighting a need for further research to improve the abilities of the models to discern speakers, especially for multilateral dialogues.

\begin{table}[t]
\centering
\begin{tabular}{@{}lccc@{}}
\toprule
\multicolumn{1}{c}{Model} & Male & Mixed & Female \\ \midrule
XGLM 7.5B & 26.8 & 27.5 & 26.2 \\
LLaMA 13B & 21.6 & 23.4 & 23.0 \\
LLaMA-2 13B & 24.9 & 25.8 & 24.9 \\
Falcon 7B & 20.5 & 22.4 & 21.6 \\
Mistral 7B & 26.9 & 26.3 & 25.5 \\
CodeLLaMA 13B & 26.8 & 26.7 & 25.4 \\
Qwen 14B & 28.7 & 27.8 & 26.6 \\
Polyglot-Ko 12.8B & 36.8 & 36.7 & 33.6 \\
KORani-v1 13B & 39.8 & 39.2 & 36.2 \\ \midrule
Human & 81.5 & 85.2 & 81.5 \\ \bottomrule
\end{tabular}
\caption{Response selection accuracy across gender compositions in bilateral dialogues. "Male" indicates both speakers are males, "Mixed" denotes dialogues between a male and female speakers, and "Female" signifies both speakers are females. Dataset: K-SNS.}
\label{tab:response-selection-gender}
\end{table}

\paragraph{Analysis on Gender Composition}
Given that gender plays a significant role in natural language processing \citep{zhao-etal-2019-gender,schofield-mehr-2016-gender}, a language model's capabilities on response selection may vary depending on the gender composition of a dialogue.
To explore this, we evaluate response selection accuracy across three types of bilateral dialogues.
As shown in Table~\ref{tab:response-selection-gender}, most models exhibit higher accuracy for male and mixed dialogues than for female dialogues.
However, human performance remains consistent across male and female dialogues, also showing higher accuracy for mixed dialogues.
As concerns around gender bias grow, it is crucial to ensure balanced progress in addressing these disparities \citep{sun-etal-2019-mitigating,liu-etal-2020-gender,kaneko-etal-2022-gender}.

\section{Conclusion}
\label{sec:conclusion}
In this study, we introduced KoDialogBench, a comprehensive benchmark tailored to evaluate Korean conversation abilities of language models.
To this end, we collected native Korean conversations from public sources or translated conversations from other languages.
Utilizing KoDialogBench, we assessed several state-of-the-art LLMs and examined how various techniques influenced their performances in Korean conversations.
Our findings emphasized the significant role of including Korean conversational data during the training phase of language models.
In addition, our results revealed that the models still lag behind human performance, highlighting an avenue for future research in developing Korean language models for conversational agents.

As the conversational capabilities of LLMs become increasingly important especially in therapeutic contexts \citep{doi:10.1080/10447318.2020.1841438,doi:10.1177/0265407520959463}, we envision KoDialogBench playing a crucial role in advancing this domain.

\section*{Limitations}
Our benchmark may suffer from a chronic problem of benchmark contamination.
Due to the scarcity of Korean language resources, there is a possibility that the held-out sources utilized to construct the benchmark might overlap with training data used for some language models.
We aim to address the detection and mitigation of benchmark contamination in our future work.

\section*{Ethics Statement}
Our benchmark dataset was designed to assess capabilities related to various situations and aspects of conversations in Korean language.
To achieve this, we utilized conversational content from publicly available datasets from various sources, either without modification or with translation if necessary.
During this process, there is a possibility that harmful content or inappropriate biases existing in the original data may have been conveyed, or may have arisen due to limitations of translation tools.
We reject any form of violence, discrimination, or offensive language, and our benchmark dataset and experimental results does not represent such values.
If any harmful content or privacy infringement is identified within the dataset, we kindly request immediate notification to the authors.
In the event of such cases being reported, we will apply the highest ethical standards and take appropriate actions.

\section*{Acknowledgements}
We would like to thank Jaehoon Lee and Sangwoo Seo for their helpful discussion.
We also appreciate the resources and support provided by Scatter Lab.
This research used datasets from `The Open AI Dataset Project (AI-Hub, S. Korea)'.
All data information can be accessed through `AI-Hub (www.aihub.or.kr)'.
This work was supported by the Institute of Information \& communications Technology Planning \& Evaluation (IITP) grant funded by the Korea government (MSIT) (No.2018-0-00584, (SW starlab) Development of Decision Support System Software based on Next-Generation Machine Learning) and No.2019-0-01906, Artificial Intelligence Graduate School Program (POSTECH)), the National Research Foundation of Korea (NRF) grant funded by the MSIT (South Korea, No.2020R1A2B5B03097210 and No. RS-2023-00217286), and the Digital Innovation Hub project supervised by the Daegu Digital Promotion Agency (DIP) grant funded by the Korean government (MSIT and Daegu Metropolitan City) in 2024 (No. DBSD1-07).

\nocite{*}
\section*{References}\label{reference}

\bibliographystyle{lrec-coling2024-natbib}
\bibliography{lrec-coling2024}

\begin{thebibliography}{73}
\expandafter\ifx\csname natexlab\endcsname\relax\def\natexlab#1{#1}\fi

\bibitem[{Ahuja et~al.(2023)Ahuja, Diddee, Hada, Ochieng, Ramesh, Jain, Nambi, Ganu, Segal, Axmed, Bali, and Sitaram}]{ahuja2023mega}
Kabir Ahuja, Harshita Diddee, Rishav Hada, Millicent Ochieng, Krithika Ramesh, Prachi Jain, Akshay Nambi, Tanuja Ganu, Sameer Segal, Maxamed Axmed, Kalika Bali, and Sunayana Sitaram. 2023.
\newblock \href {http://arxiv.org/abs/2303.12528} {Mega: Multilingual evaluation of generative ai}.

\bibitem[{Augustyniak et~al.(2022)Augustyniak, Tagowski, Sawczyn, Janiak, Bartusiak, Szymczak, Janz, Szyma\'{n}ski, W\k{a}troba, Morzy, Kajdanowicz, and Piasecki}]{NEURIPS2022_890b206e}
Lukasz Augustyniak, Kamil Tagowski, Albert Sawczyn, Denis Janiak, Roman Bartusiak, Adrian Szymczak, Arkadiusz Janz, Piotr Szyma\'{n}ski, Marcin W\k{a}troba, Miko\l~aj Morzy, Tomasz Kajdanowicz, and Maciej Piasecki. 2022.
\newblock \href {https://proceedings.neurips.cc/paper_files/paper/2022/file/890b206ebb79e550f3988cb8db936f42-Paper-Datasets_and_Benchmarks.pdf} {This is the way: designing and compiling lepiszcze, a comprehensive nlp benchmark for polish}.
\newblock In \emph{Advances in Neural Information Processing Systems}, volume~35, pages 21805--21818. Curran Associates, Inc.

\bibitem[{Bai et~al.(2023)Bai, Bai, Chu, Cui, Dang, Deng, Fan, Ge, Han, Huang, Hui, Ji, Li, Lin, Lin, Liu, Liu, Lu, Lu, Ma, Men, Ren, Ren, Tan, Tan, Tu, Wang, Wang, Wang, Wu, Xu, Xu, Yang, Yang, Yang, Yang, Yao, Yu, Yuan, Yuan, Zhang, Zhang, Zhang, Zhang, Zhou, Zhou, Zhou, and Zhu}]{bai2023qwen}
Jinze Bai, Shuai Bai, Yunfei Chu, Zeyu Cui, Kai Dang, Xiaodong Deng, Yang Fan, Wenbin Ge, Yu~Han, Fei Huang, Binyuan Hui, Luo Ji, Mei Li, Junyang Lin, Runji Lin, Dayiheng Liu, Gao Liu, Chengqiang Lu, Keming Lu, Jianxin Ma, Rui Men, Xingzhang Ren, Xuancheng Ren, Chuanqi Tan, Sinan Tan, Jianhong Tu, Peng Wang, Shijie Wang, Wei Wang, Shengguang Wu, Benfeng Xu, Jin Xu, An~Yang, Hao Yang, Jian Yang, Shusheng Yang, Yang Yao, Bowen Yu, Hongyi Yuan, Zheng Yuan, Jianwei Zhang, Xingxuan Zhang, Yichang Zhang, Zhenru Zhang, Chang Zhou, Jingren Zhou, Xiaohuan Zhou, and Tianhang Zhu. 2023.
\newblock \href {http://arxiv.org/abs/2309.16609} {Qwen technical report}.

\bibitem[{Bandarkar et~al.(2023)Bandarkar, Liang, Muller, Artetxe, Shukla, Husa, Goyal, Krishnan, Zettlemoyer, and Khabsa}]{bandarkar2023belebele}
Lucas Bandarkar, Davis Liang, Benjamin Muller, Mikel Artetxe, Satya~Narayan Shukla, Donald Husa, Naman Goyal, Abhinandan Krishnan, Luke Zettlemoyer, and Madian Khabsa. 2023.
\newblock \href {http://arxiv.org/abs/2308.16884} {The belebele benchmark: a parallel reading comprehension dataset in 122 language variants}.

\bibitem[{Bisk et~al.(2020)Bisk, Zellers, Gao, Choi et~al.}]{bisk2020piqa}
Yonatan Bisk, Rowan Zellers, Jianfeng Gao, Yejin Choi, et~al. 2020.
\newblock \href {https://ojs.aaai.org/index.php/AAAI/article/view/6239} {Piqa: Reasoning about physical commonsense in natural language}.
\newblock In \emph{Proceedings of the AAAI conference on artificial intelligence}, volume~34, pages 7432--7439.

\bibitem[{Brown et~al.(2020)Brown, Mann, Ryder, Subbiah, Kaplan, Dhariwal, Neelakantan, Shyam, Sastry, Askell, Agarwal, Herbert-Voss, Krueger, Henighan, Child, Ramesh, Ziegler, Wu, Winter, Hesse, Chen, Sigler, Litwin, Gray, Chess, Clark, Berner, McCandlish, Radford, Sutskever, and Amodei}]{NEURIPS2020_1457c0d6}
Tom Brown, Benjamin Mann, Nick Ryder, Melanie Subbiah, Jared~D Kaplan, Prafulla Dhariwal, Arvind Neelakantan, Pranav Shyam, Girish Sastry, Amanda Askell, Sandhini Agarwal, Ariel Herbert-Voss, Gretchen Krueger, Tom Henighan, Rewon Child, Aditya Ramesh, Daniel Ziegler, Jeffrey Wu, Clemens Winter, Chris Hesse, Mark Chen, Eric Sigler, Mateusz Litwin, Scott Gray, Benjamin Chess, Jack Clark, Christopher Berner, Sam McCandlish, Alec Radford, Ilya Sutskever, and Dario Amodei. 2020.
\newblock \href {https://proceedings.neurips.cc/paper_files/paper/2020/file/1457c0d6bfcb4967418bfb8ac142f64a-Paper.pdf} {Language models are few-shot learners}.
\newblock In \emph{Advances in Neural Information Processing Systems}, volume~33, pages 1877--1901. Curran Associates, Inc.

\bibitem[{Caldarini et~al.(2022)Caldarini, Jaf, and McGarry}]{caldarini2022literature}
Guendalina Caldarini, Sardar Jaf, and Kenneth McGarry. 2022.
\newblock \href {https://www.mdpi.com/2078-2489/13/1/41} {A literature survey of recent advances in chatbots}.
\newblock \emph{Information}, 13(1):41.

\bibitem[{Chatterjee et~al.(2019)Chatterjee, Narahari, Joshi, and Agrawal}]{chatterjee-etal-2019-semeval}
Ankush Chatterjee, Kedhar~Nath Narahari, Meghana Joshi, and Puneet Agrawal. 2019.
\newblock \href {https://doi.org/10.18653/v1/S19-2005} {{S}em{E}val-2019 task 3: {E}mo{C}ontext contextual emotion detection in text}.
\newblock In \emph{Proceedings of the 13th International Workshop on Semantic Evaluation}, pages 39--48, Minneapolis, Minnesota, USA. Association for Computational Linguistics.

\bibitem[{Chaves and Gerosa(2021)}]{doi:10.1080/10447318.2020.1841438}
Ana~Paula Chaves and Marco~Aurelio Gerosa. 2021.
\newblock \href {https://doi.org/10.1080/10447318.2020.1841438} {How should my chatbot interact? a survey on social characteristics in human–chatbot interaction design}.
\newblock \emph{International Journal of Human–Computer Interaction}, 37(8):729--758.

\bibitem[{Chen et~al.(2017)Chen, Liu, Yin, and Tang}]{10.1145/3166054.3166058}
Hongshen Chen, Xiaorui Liu, Dawei Yin, and Jiliang Tang. 2017.
\newblock \href {https://doi.org/10.1145/3166054.3166058} {A survey on dialogue systems: Recent advances and new frontiers}.
\newblock \emph{SIGKDD Explor. Newsl.}, 19(2):25–35.

\bibitem[{Chen et~al.(2021)Chen, Tworek, Jun, Yuan, de~Oliveira~Pinto, Kaplan, Edwards, Burda, Joseph, Brockman, Ray, Puri, Krueger, Petrov, Khlaaf, Sastry, Mishkin, Chan, Gray, Ryder, Pavlov, Power, Kaiser, Bavarian, Winter, Tillet, Such, Cummings, Plappert, Chantzis, Barnes, Herbert-Voss, Guss, Nichol, Paino, Tezak, Tang, Babuschkin, Balaji, Jain, Saunders, Hesse, Carr, Leike, Achiam, Misra, Morikawa, Radford, Knight, Brundage, Murati, Mayer, Welinder, McGrew, Amodei, McCandlish, Sutskever, and Zaremba}]{chen2021evaluating}
Mark Chen, Jerry Tworek, Heewoo Jun, Qiming Yuan, Henrique~Ponde de~Oliveira~Pinto, Jared Kaplan, Harri Edwards, Yuri Burda, Nicholas Joseph, Greg Brockman, Alex Ray, Raul Puri, Gretchen Krueger, Michael Petrov, Heidy Khlaaf, Girish Sastry, Pamela Mishkin, Brooke Chan, Scott Gray, Nick Ryder, Mikhail Pavlov, Alethea Power, Lukasz Kaiser, Mohammad Bavarian, Clemens Winter, Philippe Tillet, Felipe~Petroski Such, Dave Cummings, Matthias Plappert, Fotios Chantzis, Elizabeth Barnes, Ariel Herbert-Voss, William~Hebgen Guss, Alex Nichol, Alex Paino, Nikolas Tezak, Jie Tang, Igor Babuschkin, Suchir Balaji, Shantanu Jain, William Saunders, Christopher Hesse, Andrew~N. Carr, Jan Leike, Josh Achiam, Vedant Misra, Evan Morikawa, Alec Radford, Matthew Knight, Miles Brundage, Mira Murati, Katie Mayer, Peter Welinder, Bob McGrew, Dario Amodei, Sam McCandlish, Ilya Sutskever, and Wojciech Zaremba. 2021.
\newblock \href {http://arxiv.org/abs/2107.03374} {Evaluating large language models trained on code}.

\bibitem[{Chowdhery et~al.(2022)Chowdhery, Narang, Devlin, Bosma, Mishra, Roberts, Barham, Chung, Sutton, Gehrmann, Schuh, Shi, Tsvyashchenko, Maynez, Rao, Barnes, Tay, Shazeer, Prabhakaran, Reif, Du, Hutchinson, Pope, Bradbury, Austin, Isard, Gur-Ari, Yin, Duke, Levskaya, Ghemawat, Dev, Michalewski, Garcia, Misra, Robinson, Fedus, Zhou, Ippolito, Luan, Lim, Zoph, Spiridonov, Sepassi, Dohan, Agrawal, Omernick, Dai, Pillai, Pellat, Lewkowycz, Moreira, Child, Polozov, Lee, Zhou, Wang, Saeta, Diaz, Firat, Catasta, Wei, Meier-Hellstern, Eck, Dean, Petrov, and Fiedel}]{chowdhery2022palm}
Aakanksha Chowdhery, Sharan Narang, Jacob Devlin, Maarten Bosma, Gaurav Mishra, Adam Roberts, Paul Barham, Hyung~Won Chung, Charles Sutton, Sebastian Gehrmann, Parker Schuh, Kensen Shi, Sasha Tsvyashchenko, Joshua Maynez, Abhishek Rao, Parker Barnes, Yi~Tay, Noam Shazeer, Vinodkumar Prabhakaran, Emily Reif, Nan Du, Ben Hutchinson, Reiner Pope, James Bradbury, Jacob Austin, Michael Isard, Guy Gur-Ari, Pengcheng Yin, Toju Duke, Anselm Levskaya, Sanjay Ghemawat, Sunipa Dev, Henryk Michalewski, Xavier Garcia, Vedant Misra, Kevin Robinson, Liam Fedus, Denny Zhou, Daphne Ippolito, David Luan, Hyeontaek Lim, Barret Zoph, Alexander Spiridonov, Ryan Sepassi, David Dohan, Shivani Agrawal, Mark Omernick, Andrew~M. Dai, Thanumalayan~Sankaranarayana Pillai, Marie Pellat, Aitor Lewkowycz, Erica Moreira, Rewon Child, Oleksandr Polozov, Katherine Lee, Zongwei Zhou, Xuezhi Wang, Brennan Saeta, Mark Diaz, Orhan Firat, Michele Catasta, Jason Wei, Kathy Meier-Hellstern, Douglas Eck, Jeff Dean, Slav Petrov, and Noah Fiedel. 2022.
\newblock \href {http://arxiv.org/abs/2204.02311} {Palm: Scaling language modeling with pathways}.

\bibitem[{Chung et~al.(2022)Chung, Hou, Longpre, Zoph, Tay, Fedus, Li, Wang, Dehghani, Brahma, Webson, Gu, Dai, Suzgun, Chen, Chowdhery, Castro-Ros, Pellat, Robinson, Valter, Narang, Mishra, Yu, Zhao, Huang, Dai, Yu, Petrov, Chi, Dean, Devlin, Roberts, Zhou, Le, and Wei}]{chung2022scaling}
Hyung~Won Chung, Le~Hou, Shayne Longpre, Barret Zoph, Yi~Tay, William Fedus, Yunxuan Li, Xuezhi Wang, Mostafa Dehghani, Siddhartha Brahma, Albert Webson, Shixiang~Shane Gu, Zhuyun Dai, Mirac Suzgun, Xinyun Chen, Aakanksha Chowdhery, Alex Castro-Ros, Marie Pellat, Kevin Robinson, Dasha Valter, Sharan Narang, Gaurav Mishra, Adams Yu, Vincent Zhao, Yanping Huang, Andrew Dai, Hongkun Yu, Slav Petrov, Ed~H. Chi, Jeff Dean, Jacob Devlin, Adam Roberts, Denny Zhou, Quoc~V. Le, and Jason Wei. 2022.
\newblock \href {http://arxiv.org/abs/2210.11416} {Scaling instruction-finetuned language models}.

\bibitem[{Cobbe et~al.(2021)Cobbe, Kosaraju, Bavarian, Chen, Jun, Kaiser, Plappert, Tworek, Hilton, Nakano, Hesse, and Schulman}]{cobbe2021training}
Karl Cobbe, Vineet Kosaraju, Mohammad Bavarian, Mark Chen, Heewoo Jun, Lukasz Kaiser, Matthias Plappert, Jerry Tworek, Jacob Hilton, Reiichiro Nakano, Christopher Hesse, and John Schulman. 2021.
\newblock \href {http://arxiv.org/abs/2110.14168} {Training verifiers to solve math word problems}.

\bibitem[{Croes and Antheunis(2021)}]{doi:10.1177/0265407520959463}
Emmelyn A.~J. Croes and Marjolijn~L. Antheunis. 2021.
\newblock \href {https://doi.org/10.1177/0265407520959463} {Can we be friends with mitsuku? a longitudinal study on the process of relationship formation between humans and a social chatbot}.
\newblock \emph{Journal of Social and Personal Relationships}, 38(1):279--300.

\bibitem[{Dodge et~al.(2021)Dodge, Sap, Marasovi{\'c}, Agnew, Ilharco, Groeneveld, Mitchell, and Gardner}]{dodge-etal-2021-documenting}
Jesse Dodge, Maarten Sap, Ana Marasovi{\'c}, William Agnew, Gabriel Ilharco, Dirk Groeneveld, Margaret Mitchell, and Matt Gardner. 2021.
\newblock \href {https://doi.org/10.18653/v1/2021.emnlp-main.98} {Documenting large webtext corpora: A case study on the colossal clean crawled corpus}.
\newblock In \emph{Proceedings of the 2021 Conference on Empirical Methods in Natural Language Processing}, pages 1286--1305, Online and Punta Cana, Dominican Republic. Association for Computational Linguistics.

\bibitem[{Galley et~al.(2019)Galley, Brockett, Gao, Gao, and Dolan}]{galley2019grounded}
Michel Galley, Chris Brockett, Xiang Gao, Jianfeng Gao, and Bill Dolan. 2019.
\newblock \href {http://workshop.colips.org/dstc7/papers/DSTC7_Task_2_overview_paper.pdf} {Grounded response generation task at dstc7}.
\newblock In \emph{AAAI Dialog System Technology Challenges Workshop}.

\bibitem[{Gao et~al.(2023)Gao, Tow, Abbasi, Biderman, Black, DiPofi, Foster, Golding, Hsu, Le~Noac'h, Li, McDonell, Muennighoff, Ociepa, Phang, Reynolds, Schoelkopf, Skowron, Sutawika, Tang, Thite, Wang, Wang, and Zou}]{eval-harness}
Leo Gao, Jonathan Tow, Baber Abbasi, Stella Biderman, Sid Black, Anthony DiPofi, Charles Foster, Laurence Golding, Jeffrey Hsu, Alain Le~Noac'h, Haonan Li, Kyle McDonell, Niklas Muennighoff, Chris Ociepa, Jason Phang, Laria Reynolds, Hailey Schoelkopf, Aviya Skowron, Lintang Sutawika, Eric Tang, Anish Thite, Ben Wang, Kevin Wang, and Andy Zou. 2023.
\newblock \href {https://doi.org/10.5281/zenodo.10256836} {A framework for few-shot language model evaluation}.

\bibitem[{Guo et~al.(2018)Guo, Metallinou, Khatri, Raju, Venkatesh, and Ram}]{guo2018topicbased}
Fenfei Guo, Angeliki Metallinou, Chandra Khatri, Anirudh Raju, Anu Venkatesh, and Ashwin Ram. 2018.
\newblock \href {http://arxiv.org/abs/1801.03622} {Topic-based evaluation for conversational bots}.

\bibitem[{Ha et~al.(2020)Ha, Nam, Kang, Lee, Yang, Jung, Kim, Kim, Kim, Kim, Doh, Lee, Sung, and Kim}]{ha2020clovacall}
Jung-Woo Ha, Kihyun Nam, Jingu Kang, Sang-Woo Lee, Sohee Yang, Hyunhoon Jung, Eunmi Kim, Hyeji Kim, Soojin Kim, Hyun~Ah Kim, Kyoungtae Doh, Chan~Kyu Lee, Nako Sung, and Sunghun Kim. 2020.
\newblock \href {http://arxiv.org/abs/2004.09367} {Clovacall: Korean goal-oriented dialog speech corpus for automatic speech recognition of contact centers}.

\bibitem[{Hendrycks et~al.(2021)Hendrycks, Burns, Basart, Zou, Mazeika, Song, and Steinhardt}]{hendrycks2021measuring}
Dan Hendrycks, Collin Burns, Steven Basart, Andy Zou, Mantas Mazeika, Dawn Song, and Jacob Steinhardt. 2021.
\newblock \href {https://openreview.net/forum?id=d7KBjmI3GmQ} {Measuring massive multitask language understanding}.
\newblock In \emph{International Conference on Learning Representations}.

\bibitem[{Hsu et~al.(2018)Hsu, Chen, Kuo, Huang, and Ku}]{hsu-etal-2018-emotionlines}
Chao-Chun Hsu, Sheng-Yeh Chen, Chuan-Chun Kuo, Ting-Hao Huang, and Lun-Wei Ku. 2018.
\newblock \href {https://aclanthology.org/L18-1252} {{E}motion{L}ines: An emotion corpus of multi-party conversations}.
\newblock In \emph{Proceedings of the Eleventh International Conference on Language Resources and Evaluation ({LREC} 2018)}, Miyazaki, Japan. European Language Resources Association (ELRA).

\bibitem[{Jang et~al.(2022)Jang, Kim, Kwon, and Davis}]{jang-etal-2022-kobest}
Myeongjun Jang, Dohyung Kim, Deuk~Sin Kwon, and Eric Davis. 2022.
\newblock \href {https://aclanthology.org/2022.coling-1.325} {{K}o{BEST}: {K}orean balanced evaluation of significant tasks}.
\newblock In \emph{Proceedings of the 29th International Conference on Computational Linguistics}, pages 3697--3708, Gyeongju, Republic of Korea. International Committee on Computational Linguistics.

\bibitem[{Jia et~al.(2021)Jia, Huang, and Zhu}]{jia2021ddrel}
Qi~Jia, Hongru Huang, and Kenny~Q Zhu. 2021.
\newblock \href {https://ojs.aaai.org/index.php/AAAI/article/view/17551} {Ddrel: A new dataset for interpersonal relation classification in dyadic dialogues}.
\newblock In \emph{Proceedings of the AAAI Conference on Artificial Intelligence}, volume~35, pages 13125--13133.

\bibitem[{Jiang et~al.(2023)Jiang, Sablayrolles, Mensch, Bamford, Chaplot, de~las Casas, Bressand, Lengyel, Lample, Saulnier, Lavaud, Lachaux, Stock, Scao, Lavril, Wang, Lacroix, and Sayed}]{jiang2023mistral}
Albert~Q. Jiang, Alexandre Sablayrolles, Arthur Mensch, Chris Bamford, Devendra~Singh Chaplot, Diego de~las Casas, Florian Bressand, Gianna Lengyel, Guillaume Lample, Lucile Saulnier, Lélio~Renard Lavaud, Marie-Anne Lachaux, Pierre Stock, Teven~Le Scao, Thibaut Lavril, Thomas Wang, Timothée Lacroix, and William~El Sayed. 2023.
\newblock \href {http://arxiv.org/abs/2310.06825} {Mistral 7b}.

\bibitem[{Kaneko et~al.(2022)Kaneko, Imankulova, Bollegala, and Okazaki}]{kaneko-etal-2022-gender}
Masahiro Kaneko, Aizhan Imankulova, Danushka Bollegala, and Naoaki Okazaki. 2022.
\newblock \href {https://doi.org/10.18653/v1/2022.naacl-main.197} {Gender bias in masked language models for multiple languages}.
\newblock In \emph{Proceedings of the 2022 Conference of the North American Chapter of the Association for Computational Linguistics: Human Language Technologies}, pages 2740--2750, Seattle, United States. Association for Computational Linguistics.

\bibitem[{Ko et~al.(2023)Ko, Yang, Ryu, Choi, Yang, Hyun, Park, and Park}]{ko2023technical}
Hyunwoong Ko, Kichang Yang, Minho Ryu, Taekyoon Choi, Seungmu Yang, Jiwung Hyun, Sungho Park, and Kyubyong Park. 2023.
\newblock \href {http://arxiv.org/abs/2306.02254} {A technical report for polyglot-ko: Open-source large-scale korean language models}.

\bibitem[{Kurihara et~al.(2022)Kurihara, Kawahara, and Shibata}]{kurihara-etal-2022-jglue}
Kentaro Kurihara, Daisuke Kawahara, and Tomohide Shibata. 2022.
\newblock \href {https://aclanthology.org/2022.lrec-1.317} {{JGLUE}: {J}apanese general language understanding evaluation}.
\newblock In \emph{Proceedings of the Thirteenth Language Resources and Evaluation Conference}, pages 2957--2966, Marseille, France. European Language Resources Association.

\bibitem[{Lai et~al.(2023)Lai, Ngo, Veyseh, Man, Dernoncourt, Bui, and Nguyen}]{lai2023chatgpt}
Viet~Dac Lai, Nghia~Trung Ngo, Amir Pouran~Ben Veyseh, Hieu Man, Franck Dernoncourt, Trung Bui, and Thien~Huu Nguyen. 2023.
\newblock \href {http://arxiv.org/abs/2304.05613} {Chatgpt beyond english: Towards a comprehensive evaluation of large language models in multilingual learning}.

\bibitem[{Lee et~al.(2023{\natexlab{a}})Lee, Moon, Park, and Lim}]{lee2023datadriven}
Seugnjun Lee, Hyeonseok Moon, Chanjun Park, and Heuiseok Lim. 2023{\natexlab{a}}.
\newblock \href {http://arxiv.org/abs/2306.14514} {Data-driven approach for formality-sensitive machine translation: Language-specific handling and synthetic data generation}.

\bibitem[{Lee et~al.(2023{\natexlab{b}})Lee, Moon, Park, and Lim}]{lee-etal-2023-improving-formality}
Seungjun Lee, Hyeonseok Moon, Chanjun Park, and Heuiseok Lim. 2023{\natexlab{b}}.
\newblock \href {https://doi.org/10.18653/v1/2023.iwslt-1.40} {Improving formality-sensitive machine translation using data-centric approaches and prompt engineering}.
\newblock In \emph{Proceedings of the 20th International Conference on Spoken Language Translation (IWSLT 2023)}, pages 420--432, Toronto, Canada (in-person and online). Association for Computational Linguistics.

\bibitem[{Li et~al.(2017)Li, Su, Shen, Li, Cao, and Niu}]{li-etal-2017-dailydialog}
Yanran Li, Hui Su, Xiaoyu Shen, Wenjie Li, Ziqiang Cao, and Shuzi Niu. 2017.
\newblock \href {https://aclanthology.org/I17-1099} {{D}aily{D}ialog: A manually labelled multi-turn dialogue dataset}.
\newblock In \emph{Proceedings of the Eighth International Joint Conference on Natural Language Processing (Volume 1: Long Papers)}, pages 986--995, Taipei, Taiwan. Asian Federation of Natural Language Processing.

\bibitem[{Lin et~al.(2022)Lin, Mihaylov, Artetxe, Wang, Chen, Simig, Ott, Goyal, Bhosale, Du, Pasunuru, Shleifer, Koura, Chaudhary, O{'}Horo, Wang, Zettlemoyer, Kozareva, Diab, Stoyanov, and Li}]{lin-etal-2022-shot}
Xi~Victoria Lin, Todor Mihaylov, Mikel Artetxe, Tianlu Wang, Shuohui Chen, Daniel Simig, Myle Ott, Naman Goyal, Shruti Bhosale, Jingfei Du, Ramakanth Pasunuru, Sam Shleifer, Punit~Singh Koura, Vishrav Chaudhary, Brian O{'}Horo, Jeff Wang, Luke Zettlemoyer, Zornitsa Kozareva, Mona Diab, Veselin Stoyanov, and Xian Li. 2022.
\newblock \href {https://doi.org/10.18653/v1/2022.emnlp-main.616} {Few-shot learning with multilingual generative language models}.
\newblock In \emph{Proceedings of the 2022 Conference on Empirical Methods in Natural Language Processing}, pages 9019--9052, Abu Dhabi, United Arab Emirates. Association for Computational Linguistics.

\bibitem[{Liu et~al.(2020)Liu, Dacon, Fan, Liu, Liu, and Tang}]{liu-etal-2020-gender}
Haochen Liu, Jamell Dacon, Wenqi Fan, Hui Liu, Zitao Liu, and Jiliang Tang. 2020.
\newblock \href {https://doi.org/10.18653/v1/2020.coling-main.390} {Does gender matter? towards fairness in dialogue systems}.
\newblock In \emph{Proceedings of the 28th International Conference on Computational Linguistics}, pages 4403--4416, Barcelona, Spain (Online). International Committee on Computational Linguistics.

\bibitem[{Madaan et~al.(2022)Madaan, Zhou, Alon, Yang, and Neubig}]{madaan-etal-2022-language}
Aman Madaan, Shuyan Zhou, Uri Alon, Yiming Yang, and Graham Neubig. 2022.
\newblock \href {https://doi.org/10.18653/v1/2022.emnlp-main.90} {Language models of code are few-shot commonsense learners}.
\newblock In \emph{Proceedings of the 2022 Conference on Empirical Methods in Natural Language Processing}, pages 1384--1403, Abu Dhabi, United Arab Emirates. Association for Computational Linguistics.

\bibitem[{Magar and Schwartz(2022)}]{magar-schwartz-2022-data}
Inbal Magar and Roy Schwartz. 2022.
\newblock \href {https://doi.org/10.18653/v1/2022.acl-short.18} {Data contamination: From memorization to exploitation}.
\newblock In \emph{Proceedings of the 60th Annual Meeting of the Association for Computational Linguistics (Volume 2: Short Papers)}, pages 157--165, Dublin, Ireland. Association for Computational Linguistics.

\bibitem[{Maynez et~al.(2023)Maynez, Agrawal, and Gehrmann}]{maynez-etal-2023-benchmarking}
Joshua Maynez, Priyanka Agrawal, and Sebastian Gehrmann. 2023.
\newblock \href {https://doi.org/10.18653/v1/2023.acl-long.511} {Benchmarking large language model capabilities for conditional generation}.
\newblock In \emph{Proceedings of the 61st Annual Meeting of the Association for Computational Linguistics (Volume 1: Long Papers)}, pages 9194--9213, Toronto, Canada. Association for Computational Linguistics.

\bibitem[{Mehri et~al.(2020)Mehri, Eric, and Hakkani-Tur}]{mehri2020dialoglue}
Shikib Mehri, Mihail Eric, and Dilek Hakkani-Tur. 2020.
\newblock \href {http://arxiv.org/abs/2009.13570} {Dialoglue: A natural language understanding benchmark for task-oriented dialogue}.

\bibitem[{Ni et~al.(2022)Ni, Young, Pandelea, Xue, and Cambria}]{10.1007/s10462-022-10248-8}
Jinjie Ni, Tom Young, Vlad Pandelea, Fuzhao Xue, and Erik Cambria. 2022.
\newblock \href {https://doi.org/10.1007/s10462-022-10248-8} {Recent advances in deep learning based dialogue systems: A systematic survey}.
\newblock \emph{Artif. Intell. Rev.}, 56(4):3055–3155.

\bibitem[{OpenAI(2023)}]{openai2023gpt4}
OpenAI. 2023.
\newblock \href {http://arxiv.org/abs/2303.08774} {Gpt-4 technical report}.

\bibitem[{Park et~al.(2021)Park, Moon, Kim, Cho, Han, Park, Song, Kim, Song, Oh, Lee, Oh, Lyu, Jeong, Lee, Seo, Lee, Kim, Lee, Jang, Do, Kim, Lim, Lee, Park, Shin, Kim, Park, Park, Oh, Ha~(NAVER AI~Lab), Cho, and Cho}]{NEURIPS2021_98dce83d}
Sungjoon Park, Jihyung Moon, Sungdong Kim, Won~Ik Cho, Ji~Yoon Han, Jangwon Park, Chisung Song, Junseong Kim, Youngsook Song, Taehwan Oh, Joohong Lee, Juhyun Oh, Sungwon Lyu, Younghoon Jeong, Inkwon Lee, Sangwoo Seo, Dongjun Lee, Hyunwoo Kim, Myeonghwa Lee, Seongbo Jang, Seungwon Do, Sunkyoung Kim, Kyungtae Lim, Jongwon Lee, Kyumin Park, Jamin Shin, Seonghyun Kim, Lucy Park, Lucy Park, Alice Oh, Jung-Woo Ha~(NAVER AI~Lab), Kyunghyun Cho, and Kyunghyun Cho. 2021.
\newblock \href {https://datasets-benchmarks-proceedings.neurips.cc/paper_files/paper/2021/file/98dce83da57b0395e163467c9dae521b-Paper-round2.pdf} {Klue: Korean language understanding evaluation}.
\newblock In \emph{Proceedings of the Neural Information Processing Systems Track on Datasets and Benchmarks}, volume~1. Curran.

\bibitem[{Penedo et~al.(2023)Penedo, Malartic, Hesslow, Cojocaru, Cappelli, Alobeidli, Pannier, Almazrouei, and Launay}]{penedo2023refinedweb}
Guilherme Penedo, Quentin Malartic, Daniel Hesslow, Ruxandra Cojocaru, Alessandro Cappelli, Hamza Alobeidli, Baptiste Pannier, Ebtesam Almazrouei, and Julien Launay. 2023.
\newblock \href {http://arxiv.org/abs/2306.01116} {The refinedweb dataset for falcon llm: Outperforming curated corpora with web data, and web data only}.

\bibitem[{Poria et~al.(2019)Poria, Hazarika, Majumder, Naik, Cambria, and Mihalcea}]{poria-etal-2019-meld}
Soujanya Poria, Devamanyu Hazarika, Navonil Majumder, Gautam Naik, Erik Cambria, and Rada Mihalcea. 2019.
\newblock \href {https://doi.org/10.18653/v1/P19-1050} {{MELD}: A multimodal multi-party dataset for emotion recognition in conversations}.
\newblock In \emph{Proceedings of the 57th Annual Meeting of the Association for Computational Linguistics}, pages 527--536, Florence, Italy. Association for Computational Linguistics.

\bibitem[{Rashkin et~al.(2019)Rashkin, Smith, Li, and Boureau}]{rashkin-etal-2019-towards}
Hannah Rashkin, Eric~Michael Smith, Margaret Li, and Y-Lan Boureau. 2019.
\newblock \href {https://doi.org/10.18653/v1/P19-1534} {Towards empathetic open-domain conversation models: A new benchmark and dataset}.
\newblock In \emph{Proceedings of the 57th Annual Meeting of the Association for Computational Linguistics}, pages 5370--5381, Florence, Italy. Association for Computational Linguistics.

\bibitem[{Rozière et~al.(2023)Rozière, Gehring, Gloeckle, Sootla, Gat, Tan, Adi, Liu, Remez, Rapin, Kozhevnikov, Evtimov, Bitton, Bhatt, Ferrer, Grattafiori, Xiong, Défossez, Copet, Azhar, Touvron, Martin, Usunier, Scialom, and Synnaeve}]{roziere2023code}
Baptiste Rozière, Jonas Gehring, Fabian Gloeckle, Sten Sootla, Itai Gat, Xiaoqing~Ellen Tan, Yossi Adi, Jingyu Liu, Tal Remez, Jérémy Rapin, Artyom Kozhevnikov, Ivan Evtimov, Joanna Bitton, Manish Bhatt, Cristian~Canton Ferrer, Aaron Grattafiori, Wenhan Xiong, Alexandre Défossez, Jade Copet, Faisal Azhar, Hugo Touvron, Louis Martin, Nicolas Usunier, Thomas Scialom, and Gabriel Synnaeve. 2023.
\newblock \href {http://arxiv.org/abs/2308.12950} {Code llama: Open foundation models for code}.

\bibitem[{Ryan et~al.(2023)Ryan, Naous, and Xu}]{ryan-etal-2023-revisiting}
Michael Ryan, Tarek Naous, and Wei Xu. 2023.
\newblock \href {https://doi.org/10.18653/v1/2023.acl-long.269} {Revisiting non-{E}nglish text simplification: A unified multilingual benchmark}.
\newblock In \emph{Proceedings of the 61st Annual Meeting of the Association for Computational Linguistics (Volume 1: Long Papers)}, pages 4898--4927, Toronto, Canada. Association for Computational Linguistics.

\bibitem[{Sap et~al.(2022)Sap, Le~Bras, Fried, and Choi}]{sap-etal-2022-neural}
Maarten Sap, Ronan Le~Bras, Daniel Fried, and Yejin Choi. 2022.
\newblock \href {https://doi.org/10.18653/v1/2022.emnlp-main.248} {Neural theory-of-mind? on the limits of social intelligence in large {LM}s}.
\newblock In \emph{Proceedings of the 2022 Conference on Empirical Methods in Natural Language Processing}, pages 3762--3780, Abu Dhabi, United Arab Emirates. Association for Computational Linguistics.

\bibitem[{Sap et~al.(2019)Sap, Rashkin, Chen, Le~Bras, and Choi}]{sap-etal-2019-social}
Maarten Sap, Hannah Rashkin, Derek Chen, Ronan Le~Bras, and Yejin Choi. 2019.
\newblock \href {https://doi.org/10.18653/v1/D19-1454} {Social {IQ}a: Commonsense reasoning about social interactions}.
\newblock In \emph{Proceedings of the 2019 Conference on Empirical Methods in Natural Language Processing and the 9th International Joint Conference on Natural Language Processing (EMNLP-IJCNLP)}, pages 4463--4473, Hong Kong, China. Association for Computational Linguistics.

\bibitem[{Schofield and Mehr(2016)}]{schofield-mehr-2016-gender}
Alexandra Schofield and Leo Mehr. 2016.
\newblock \href {https://doi.org/10.18653/v1/W16-0204} {Gender-distinguishing features in film dialogue}.
\newblock In \emph{Proceedings of the Fifth Workshop on Computational Linguistics for Literature}, pages 32--39, San Diego, California, USA. Association for Computational Linguistics.

\bibitem[{Shriberg et~al.(2004)Shriberg, Dhillon, Bhagat, Ang, and Carvey}]{shriberg-etal-2004-icsi}
Elizabeth Shriberg, Raj Dhillon, Sonali Bhagat, Jeremy Ang, and Hannah Carvey. 2004.
\newblock \href {https://aclanthology.org/W04-2319} {The {ICSI} meeting recorder dialog act ({MRDA}) corpus}.
\newblock In \emph{Proceedings of the 5th {SIG}dial Workshop on Discourse and Dialogue at {HLT}-{NAACL} 2004}, pages 97--100, Cambridge, Massachusetts, USA. Association for Computational Linguistics.

\bibitem[{Shuster et~al.(2020)Shuster, Ju, Roller, Dinan, Boureau, and Weston}]{shuster-etal-2020-dialogue}
Kurt Shuster, Da~Ju, Stephen Roller, Emily Dinan, Y-Lan Boureau, and Jason Weston. 2020.
\newblock \href {https://doi.org/10.18653/v1/2020.acl-main.222} {The dialogue dodecathlon: Open-domain knowledge and image grounded conversational agents}.
\newblock In \emph{Proceedings of the 58th Annual Meeting of the Association for Computational Linguistics}, pages 2453--2470, Online. Association for Computational Linguistics.

\bibitem[{Srivastava et~al.(2023)Srivastava, Rastogi, Rao, Shoeb, Abid, Fisch, Brown, Santoro, Gupta, Garriga-Alonso, Kluska, Lewkowycz, Agarwal, Power, Ray, Warstadt, Kocurek, Safaya, Tazarv, Xiang, Parrish, Nie, Hussain, Askell, Dsouza, Slone, Rahane, Iyer, Andreassen, Madotto, Santilli, Stuhlm{\"u}ller, Dai, La, Lampinen, Zou, Jiang, Chen, Vuong, Gupta, Gottardi, Norelli, Venkatesh, Gholamidavoodi, Tabassum, Menezes, Kirubarajan, Mullokandov, Sabharwal, Herrick, Efrat, Erdem, Karaka{\c{s}}, Roberts, Loe, Zoph, Bojanowski, {\"O}zyurt, Hedayatnia, Neyshabur, Inden, Stein, Ekmekci, Lin, Howald, Orinion, Diao, Dour, Stinson, Argueta, Ferri, Singh, Rathkopf, Meng, Baral, Wu, Callison-Burch, Waites, Voigt, Manning, Potts, Ramirez, Rivera, Siro, Raffel, Ashcraft, Garbacea, Sileo, Garrette, Hendrycks, Kilman, Roth, Freeman, Khashabi, Levy, Gonz{\'a}lez, Perszyk, Hernandez, Chen, Ippolito, Gilboa, Dohan, Drakard, Jurgens, Datta, Ganguli, Emelin, Kleyko, Yuret, Chen, Tam, Hupkes, Misra, Buzan, Mollo, Yang, Lee,
  Schrader, Shutova, Cubuk, Segal, Hagerman, Barnes, Donoway, Pavlick, Rodol{\`a}, Lam, Chu, Tang, Erdem, Chang, Chi, Dyer, Jerzak, Kim, Manyasi, Zheltonozhskii, Xia, Siar, Mart{\'\i}nez-Plumed, Happ{\'e}, Chollet, Rong, Mishra, Winata, de~Melo, Kruszewski, Parascandolo, Mariani, Wang, Jaimovitch-Lopez, Betz, Gur-Ari, Galijasevic, Kim, Rashkin, Hajishirzi, Mehta, Bogar, Shevlin, Schuetze, Yakura, Zhang, Wong, Ng, Noble, Jumelet, Geissinger, Kernion, Hilton, Lee, Fisac, Simon, Koppel, Zheng, Zou, Kocon, Thompson, Wingfield, Kaplan, Radom, Sohl-Dickstein, Phang, Wei, Yosinski, Novikova, Bosscher, Marsh, Kim, Taal, Engel, Alabi, Xu, Song, Tang, Waweru, Burden, Miller, Balis, Batchelder, Berant, Frohberg, Rozen, Hernandez-Orallo, Boudeman, Guerr, Jones, Tenenbaum, Rule, Chua, Kanclerz, Livescu, Krauth, Gopalakrishnan, Ignatyeva, Markert, Dhole, Gimpel, Omondi, Mathewson, Chiafullo, Shkaruta, Shridhar, McDonell, Richardson, Reynolds, Gao, Zhang, Dugan, Qin, Contreras-Ochando, Morency, Moschella, Lam, Noble,
  Schmidt, He, Oliveros-Col{\'o}n, Metz, Senel, Bosma, Sap, Hoeve, Farooqi, Faruqui, Mazeika, Baturan, Marelli, Maru, Ramirez-Quintana, Tolkiehn, Giulianelli, Lewis, Potthast, Leavitt, Hagen, Schubert, Baitemirova, Arnaud, McElrath, Yee, Cohen, Gu, Ivanitskiy, Starritt, Strube, Sw{\k{e}}drowski, Bevilacqua, Yasunaga, Kale, Cain, Xu, Suzgun, Walker, Tiwari, Bansal, Aminnaseri, Geva, Gheini, T, Peng, Chi, Lee, Krakover, Cameron, Roberts, Doiron, Martinez, Nangia, Deckers, Muennighoff, Keskar, Iyer, Constant, Fiedel, Wen, Zhang, Agha, Elbaghdadi, Levy, Evans, Casares, Doshi, Fung, Liang, Vicol, Alipoormolabashi, Liao, Liang, Chang, Eckersley, Htut, Hwang, Mi{\l}kowski, Patil, Pezeshkpour, Oli, Mei, Lyu, Chen, Banjade, Rudolph, Gabriel, Habacker, Risco, Milli{\`e}re, Garg, Barnes, Saurous, Arakawa, Raymaekers, Frank, Sikand, Novak, Sitelew, Bras, Liu, Jacobs, Zhang, Salakhutdinov, Chi, Lee, Stovall, Teehan, Yang, Singh, Mohammad, Anand, Dillavou, Shleifer, Wiseman, Gruetter, Bowman, Schoenholz, Han, Kwatra, Rous,
  Ghazarian, Ghosh, Casey, Bischoff, Gehrmann, Schuster, Sadeghi, Hamdan, Zhou, Srivastava, Shi, Singh, Asaadi, Gu, Pachchigar, Toshniwal, Upadhyay, Debnath, Shakeri, Thormeyer, Melzi, Reddy, Makini, Lee, Torene, Hatwar, Dehaene, Divic, Ermon, Biderman, Lin, Prasad, Piantadosi, Shieber, Misherghi, Kiritchenko, Mishra, Linzen, Schuster, Li, Yu, Ali, Hashimoto, Wu, Desbordes, Rothschild, Phan, Wang, Nkinyili, Schick, Kornev, Tunduny, Gerstenberg, Chang, Neeraj, Khot, Shultz, Shaham, Misra, Demberg, Nyamai, Raunak, Ramasesh, vinay~uday prabhu, Padmakumar, Srikumar, Fedus, Saunders, Zhang, Vossen, Ren, Tong, Zhao, Wu, Shen, Yaghoobzadeh, Lakretz, Song, Bahri, Choi, Yang, Hao, Chen, Belinkov, Hou, Hou, Bai, Seid, Zhao, Wang, Wang, Wang, and Wu}]{srivastava2023beyond}
Aarohi Srivastava, Abhinav Rastogi, Abhishek Rao, Abu Awal~Md Shoeb, Abubakar Abid, Adam Fisch, Adam~R. Brown, Adam Santoro, Aditya Gupta, Adri{\`a} Garriga-Alonso, Agnieszka Kluska, Aitor Lewkowycz, Akshat Agarwal, Alethea Power, Alex Ray, Alex Warstadt, Alexander~W. Kocurek, Ali Safaya, Ali Tazarv, Alice Xiang, Alicia Parrish, Allen Nie, Aman Hussain, Amanda Askell, Amanda Dsouza, Ambrose Slone, Ameet Rahane, Anantharaman~S. Iyer, Anders~Johan Andreassen, Andrea Madotto, Andrea Santilli, Andreas Stuhlm{\"u}ller, Andrew~M. Dai, Andrew La, Andrew Lampinen, Andy Zou, Angela Jiang, Angelica Chen, Anh Vuong, Animesh Gupta, Anna Gottardi, Antonio Norelli, Anu Venkatesh, Arash Gholamidavoodi, Arfa Tabassum, Arul Menezes, Arun Kirubarajan, Asher Mullokandov, Ashish Sabharwal, Austin Herrick, Avia Efrat, Aykut Erdem, Ayla Karaka{\c{s}}, B.~Ryan Roberts, Bao~Sheng Loe, Barret Zoph, Bart{\l}omiej Bojanowski, Batuhan {\"O}zyurt, Behnam Hedayatnia, Behnam Neyshabur, Benjamin Inden, Benno Stein, Berk Ekmekci, Bill~Yuchen
  Lin, Blake Howald, Bryan Orinion, Cameron Diao, Cameron Dour, Catherine Stinson, Cedrick Argueta, Cesar Ferri, Chandan Singh, Charles Rathkopf, Chenlin Meng, Chitta Baral, Chiyu Wu, Chris Callison-Burch, Christopher Waites, Christian Voigt, Christopher~D Manning, Christopher Potts, Cindy Ramirez, Clara~E. Rivera, Clemencia Siro, Colin Raffel, Courtney Ashcraft, Cristina Garbacea, Damien Sileo, Dan Garrette, Dan Hendrycks, Dan Kilman, Dan Roth, C.~Daniel Freeman, Daniel Khashabi, Daniel Levy, Daniel~Mosegu{\'\i} Gonz{\'a}lez, Danielle Perszyk, Danny Hernandez, Danqi Chen, Daphne Ippolito, Dar Gilboa, David Dohan, David Drakard, David Jurgens, Debajyoti Datta, Deep Ganguli, Denis Emelin, Denis Kleyko, Deniz Yuret, Derek Chen, Derek Tam, Dieuwke Hupkes, Diganta Misra, Dilyar Buzan, Dimitri~Coelho Mollo, Diyi Yang, Dong-Ho Lee, Dylan Schrader, Ekaterina Shutova, Ekin~Dogus Cubuk, Elad Segal, Eleanor Hagerman, Elizabeth Barnes, Elizabeth Donoway, Ellie Pavlick, Emanuele Rodol{\`a}, Emma Lam, Eric Chu, Eric Tang,
  Erkut Erdem, Ernie Chang, Ethan~A Chi, Ethan Dyer, Ethan Jerzak, Ethan Kim, Eunice~Engefu Manyasi, Evgenii Zheltonozhskii, Fanyue Xia, Fatemeh Siar, Fernando Mart{\'\i}nez-Plumed, Francesca Happ{\'e}, Francois Chollet, Frieda Rong, Gaurav Mishra, Genta~Indra Winata, Gerard de~Melo, Germ{\'a}n Kruszewski, Giambattista Parascandolo, Giorgio Mariani, Gloria~Xinyue Wang, Gonzalo Jaimovitch-Lopez, Gregor Betz, Guy Gur-Ari, Hana Galijasevic, Hannah Kim, Hannah Rashkin, Hannaneh Hajishirzi, Harsh Mehta, Hayden Bogar, Henry Francis~Anthony Shevlin, Hinrich Schuetze, Hiromu Yakura, Hongming Zhang, Hugh~Mee Wong, Ian Ng, Isaac Noble, Jaap Jumelet, Jack Geissinger, Jackson Kernion, Jacob Hilton, Jaehoon Lee, Jaime~Fern{\'a}ndez Fisac, James~B Simon, James Koppel, James Zheng, James Zou, Jan Kocon, Jana Thompson, Janelle Wingfield, Jared Kaplan, Jarema Radom, Jascha Sohl-Dickstein, Jason Phang, Jason Wei, Jason Yosinski, Jekaterina Novikova, Jelle Bosscher, Jennifer Marsh, Jeremy Kim, Jeroen Taal, Jesse Engel, Jesujoba
  Alabi, Jiacheng Xu, Jiaming Song, Jillian Tang, Joan Waweru, John Burden, John Miller, John~U. Balis, Jonathan Batchelder, Jonathan Berant, J{\"o}rg Frohberg, Jos Rozen, Jose Hernandez-Orallo, Joseph Boudeman, Joseph Guerr, Joseph Jones, Joshua~B. Tenenbaum, Joshua~S. Rule, Joyce Chua, Kamil Kanclerz, Karen Livescu, Karl Krauth, Karthik Gopalakrishnan, Katerina Ignatyeva, Katja Markert, Kaustubh Dhole, Kevin Gimpel, Kevin Omondi, Kory~Wallace Mathewson, Kristen Chiafullo, Ksenia Shkaruta, Kumar Shridhar, Kyle McDonell, Kyle Richardson, Laria Reynolds, Leo Gao, Li~Zhang, Liam Dugan, Lianhui Qin, Lidia Contreras-Ochando, Louis-Philippe Morency, Luca Moschella, Lucas Lam, Lucy Noble, Ludwig Schmidt, Luheng He, Luis Oliveros-Col{\'o}n, Luke Metz, L{\"u}tfi~Kerem Senel, Maarten Bosma, Maarten Sap, Maartje~Ter Hoeve, Maheen Farooqi, Manaal Faruqui, Mantas Mazeika, Marco Baturan, Marco Marelli, Marco Maru, Maria~Jose Ramirez-Quintana, Marie Tolkiehn, Mario Giulianelli, Martha Lewis, Martin Potthast, Matthew~L
  Leavitt, Matthias Hagen, M{\'a}ty{\'a}s Schubert, Medina~Orduna Baitemirova, Melody Arnaud, Melvin McElrath, Michael~Andrew Yee, Michael Cohen, Michael Gu, Michael Ivanitskiy, Michael Starritt, Michael Strube, Micha{\l} Sw{\k{e}}drowski, Michele Bevilacqua, Michihiro Yasunaga, Mihir Kale, Mike Cain, Mimee Xu, Mirac Suzgun, Mitch Walker, Mo~Tiwari, Mohit Bansal, Moin Aminnaseri, Mor Geva, Mozhdeh Gheini, Mukund~Varma T, Nanyun Peng, Nathan~Andrew Chi, Nayeon Lee, Neta Gur-Ari Krakover, Nicholas Cameron, Nicholas Roberts, Nick Doiron, Nicole Martinez, Nikita Nangia, Niklas Deckers, Niklas Muennighoff, Nitish~Shirish Keskar, Niveditha~S. Iyer, Noah Constant, Noah Fiedel, Nuan Wen, Oliver Zhang, Omar Agha, Omar Elbaghdadi, Omer Levy, Owain Evans, Pablo Antonio~Moreno Casares, Parth Doshi, Pascale Fung, Paul~Pu Liang, Paul Vicol, Pegah Alipoormolabashi, Peiyuan Liao, Percy Liang, Peter~W Chang, Peter Eckersley, Phu~Mon Htut, Pinyu Hwang, Piotr Mi{\l}kowski, Piyush Patil, Pouya Pezeshkpour, Priti Oli, Qiaozhu
  Mei, Qing Lyu, Qinlang Chen, Rabin Banjade, Rachel~Etta Rudolph, Raefer Gabriel, Rahel Habacker, Ramon Risco, Rapha{\"e}l Milli{\`e}re, Rhythm Garg, Richard Barnes, Rif~A. Saurous, Riku Arakawa, Robbe Raymaekers, Robert Frank, Rohan Sikand, Roman Novak, Roman Sitelew, Ronan~Le Bras, Rosanne Liu, Rowan Jacobs, Rui Zhang, Russ Salakhutdinov, Ryan~Andrew Chi, Seungjae~Ryan Lee, Ryan Stovall, Ryan Teehan, Rylan Yang, Sahib Singh, Saif~M. Mohammad, Sajant Anand, Sam Dillavou, Sam Shleifer, Sam Wiseman, Samuel Gruetter, Samuel~R. Bowman, Samuel~Stern Schoenholz, Sanghyun Han, Sanjeev Kwatra, Sarah~A. Rous, Sarik Ghazarian, Sayan Ghosh, Sean Casey, Sebastian Bischoff, Sebastian Gehrmann, Sebastian Schuster, Sepideh Sadeghi, Shadi Hamdan, Sharon Zhou, Shashank Srivastava, Sherry Shi, Shikhar Singh, Shima Asaadi, Shixiang~Shane Gu, Shubh Pachchigar, Shubham Toshniwal, Shyam Upadhyay, Shyamolima~Shammie Debnath, Siamak Shakeri, Simon Thormeyer, Simone Melzi, Siva Reddy, Sneha~Priscilla Makini, Soo-Hwan Lee, Spencer
  Torene, Sriharsha Hatwar, Stanislas Dehaene, Stefan Divic, Stefano Ermon, Stella Biderman, Stephanie Lin, Stephen Prasad, Steven Piantadosi, Stuart Shieber, Summer Misherghi, Svetlana Kiritchenko, Swaroop Mishra, Tal Linzen, Tal Schuster, Tao Li, Tao Yu, Tariq Ali, Tatsunori Hashimoto, Te-Lin Wu, Th{\'e}o Desbordes, Theodore Rothschild, Thomas Phan, Tianle Wang, Tiberius Nkinyili, Timo Schick, Timofei Kornev, Titus Tunduny, Tobias Gerstenberg, Trenton Chang, Trishala Neeraj, Tushar Khot, Tyler Shultz, Uri Shaham, Vedant Misra, Vera Demberg, Victoria Nyamai, Vikas Raunak, Vinay~Venkatesh Ramasesh, vinay~uday prabhu, Vishakh Padmakumar, Vivek Srikumar, William Fedus, William Saunders, William Zhang, Wout Vossen, Xiang Ren, Xiaoyu Tong, Xinran Zhao, Xinyi Wu, Xudong Shen, Yadollah Yaghoobzadeh, Yair Lakretz, Yangqiu Song, Yasaman Bahri, Yejin Choi, Yichi Yang, Yiding Hao, Yifu Chen, Yonatan Belinkov, Yu~Hou, Yufang Hou, Yuntao Bai, Zachary Seid, Zhuoye Zhao, Zijian Wang, Zijie~J. Wang, Zirui Wang, and Ziyi Wu.
  2023.
\newblock \href {https://openreview.net/forum?id=uyTL5Bvosj} {Beyond the imitation game: Quantifying and extrapolating the capabilities of language models}.
\newblock \emph{Transactions on Machine Learning Research}.

\bibitem[{Stolcke et~al.(2000)Stolcke, Ries, Coccaro, Shriberg, Bates, Jurafsky, Taylor, Martin, Van Ess-Dykema, and Meteer}]{stolcke-etal-2000-dialogue}
Andreas Stolcke, Klaus Ries, Noah Coccaro, Elizabeth Shriberg, Rebecca Bates, Daniel Jurafsky, Paul Taylor, Rachel Martin, Carol Van Ess-Dykema, and Marie Meteer. 2000.
\newblock \href {https://aclanthology.org/J00-3003} {Dialogue act modeling for automatic tagging and recognition of conversational speech}.
\newblock \emph{Computational Linguistics}, 26(3):339--374.

\bibitem[{Sun et~al.(2019)Sun, Gaut, Tang, Huang, ElSherief, Zhao, Mirza, Belding, Chang, and Wang}]{sun-etal-2019-mitigating}
Tony Sun, Andrew Gaut, Shirlyn Tang, Yuxin Huang, Mai ElSherief, Jieyu Zhao, Diba Mirza, Elizabeth Belding, Kai-Wei Chang, and William~Yang Wang. 2019.
\newblock \href {https://doi.org/10.18653/v1/P19-1159} {Mitigating gender bias in natural language processing: Literature review}.
\newblock In \emph{Proceedings of the 57th Annual Meeting of the Association for Computational Linguistics}, pages 1630--1640, Florence, Italy. Association for Computational Linguistics.

\bibitem[{Suzgun et~al.(2023)Suzgun, Scales, Sch{\"a}rli, Gehrmann, Tay, Chung, Chowdhery, Le, Chi, Zhou, and Wei}]{suzgun-etal-2023-challenging}
Mirac Suzgun, Nathan Scales, Nathanael Sch{\"a}rli, Sebastian Gehrmann, Yi~Tay, Hyung~Won Chung, Aakanksha Chowdhery, Quoc Le, Ed~Chi, Denny Zhou, and Jason Wei. 2023.
\newblock \href {https://doi.org/10.18653/v1/2023.findings-acl.824} {Challenging {BIG}-bench tasks and whether chain-of-thought can solve them}.
\newblock In \emph{Findings of the Association for Computational Linguistics: ACL 2023}, pages 13003--13051, Toronto, Canada. Association for Computational Linguistics.

\bibitem[{Taori et~al.(2023)Taori, Gulrajani, Zhang, Dubois, Li, Guestrin, Liang, and Hashimoto}]{alpaca}
Rohan Taori, Ishaan Gulrajani, Tianyi Zhang, Yann Dubois, Xuechen Li, Carlos Guestrin, Percy Liang, and Tatsunori~B. Hashimoto. 2023.
\newblock \href {https://github.com/tatsu-lab/stanford_alpaca} {Stanford alpaca: An instruction-following llama model}.

\bibitem[{Thoppilan et~al.(2022)Thoppilan, Freitas, Hall, Shazeer, Kulshreshtha, Cheng, Jin, Bos, Baker, Du, Li, Lee, Zheng, Ghafouri, Menegali, Huang, Krikun, Lepikhin, Qin, Chen, Xu, Chen, Roberts, Bosma, Zhao, Zhou, Chang, Krivokon, Rusch, Pickett, Srinivasan, Man, Meier-Hellstern, Morris, Doshi, Santos, Duke, Soraker, Zevenbergen, Prabhakaran, Diaz, Hutchinson, Olson, Molina, Hoffman-John, Lee, Aroyo, Rajakumar, Butryna, Lamm, Kuzmina, Fenton, Cohen, Bernstein, Kurzweil, Aguera-Arcas, Cui, Croak, Chi, and Le}]{thoppilan2022lamda}
Romal Thoppilan, Daniel~De Freitas, Jamie Hall, Noam Shazeer, Apoorv Kulshreshtha, Heng-Tze Cheng, Alicia Jin, Taylor Bos, Leslie Baker, Yu~Du, YaGuang Li, Hongrae Lee, Huaixiu~Steven Zheng, Amin Ghafouri, Marcelo Menegali, Yanping Huang, Maxim Krikun, Dmitry Lepikhin, James Qin, Dehao Chen, Yuanzhong Xu, Zhifeng Chen, Adam Roberts, Maarten Bosma, Vincent Zhao, Yanqi Zhou, Chung-Ching Chang, Igor Krivokon, Will Rusch, Marc Pickett, Pranesh Srinivasan, Laichee Man, Kathleen Meier-Hellstern, Meredith~Ringel Morris, Tulsee Doshi, Renelito~Delos Santos, Toju Duke, Johnny Soraker, Ben Zevenbergen, Vinodkumar Prabhakaran, Mark Diaz, Ben Hutchinson, Kristen Olson, Alejandra Molina, Erin Hoffman-John, Josh Lee, Lora Aroyo, Ravi Rajakumar, Alena Butryna, Matthew Lamm, Viktoriya Kuzmina, Joe Fenton, Aaron Cohen, Rachel Bernstein, Ray Kurzweil, Blaise Aguera-Arcas, Claire Cui, Marian Croak, Ed~Chi, and Quoc Le. 2022.
\newblock \href {http://arxiv.org/abs/2201.08239} {Lamda: Language models for dialog applications}.

\bibitem[{Touvron et~al.(2023{\natexlab{a}})Touvron, Lavril, Izacard, Martinet, Lachaux, Lacroix, Rozière, Goyal, Hambro, Azhar, Rodriguez, Joulin, Grave, and Lample}]{touvron2023llama}
Hugo Touvron, Thibaut Lavril, Gautier Izacard, Xavier Martinet, Marie-Anne Lachaux, Timothée Lacroix, Baptiste Rozière, Naman Goyal, Eric Hambro, Faisal Azhar, Aurelien Rodriguez, Armand Joulin, Edouard Grave, and Guillaume Lample. 2023{\natexlab{a}}.
\newblock \href {http://arxiv.org/abs/2302.13971} {Llama: Open and efficient foundation language models}.

\bibitem[{Touvron et~al.(2023{\natexlab{b}})Touvron, Martin, Stone, Albert, Almahairi, Babaei, Bashlykov, Batra, Bhargava, Bhosale, Bikel, Blecher, Ferrer, Chen, Cucurull, Esiobu, Fernandes, Fu, Fu, Fuller, Gao, Goswami, Goyal, Hartshorn, Hosseini, Hou, Inan, Kardas, Kerkez, Khabsa, Kloumann, Korenev, Koura, Lachaux, Lavril, Lee, Liskovich, Lu, Mao, Martinet, Mihaylov, Mishra, Molybog, Nie, Poulton, Reizenstein, Rungta, Saladi, Schelten, Silva, Smith, Subramanian, Tan, Tang, Taylor, Williams, Kuan, Xu, Yan, Zarov, Zhang, Fan, Kambadur, Narang, Rodriguez, Stojnic, Edunov, and Scialom}]{touvron2023llama2}
Hugo Touvron, Louis Martin, Kevin Stone, Peter Albert, Amjad Almahairi, Yasmine Babaei, Nikolay Bashlykov, Soumya Batra, Prajjwal Bhargava, Shruti Bhosale, Dan Bikel, Lukas Blecher, Cristian~Canton Ferrer, Moya Chen, Guillem Cucurull, David Esiobu, Jude Fernandes, Jeremy Fu, Wenyin Fu, Brian Fuller, Cynthia Gao, Vedanuj Goswami, Naman Goyal, Anthony Hartshorn, Saghar Hosseini, Rui Hou, Hakan Inan, Marcin Kardas, Viktor Kerkez, Madian Khabsa, Isabel Kloumann, Artem Korenev, Punit~Singh Koura, Marie-Anne Lachaux, Thibaut Lavril, Jenya Lee, Diana Liskovich, Yinghai Lu, Yuning Mao, Xavier Martinet, Todor Mihaylov, Pushkar Mishra, Igor Molybog, Yixin Nie, Andrew Poulton, Jeremy Reizenstein, Rashi Rungta, Kalyan Saladi, Alan Schelten, Ruan Silva, Eric~Michael Smith, Ranjan Subramanian, Xiaoqing~Ellen Tan, Binh Tang, Ross Taylor, Adina Williams, Jian~Xiang Kuan, Puxin Xu, Zheng Yan, Iliyan Zarov, Yuchen Zhang, Angela Fan, Melanie Kambadur, Sharan Narang, Aurelien Rodriguez, Robert Stojnic, Sergey Edunov, and Thomas
  Scialom. 2023{\natexlab{b}}.
\newblock \href {http://arxiv.org/abs/2307.09288} {Llama 2: Open foundation and fine-tuned chat models}.

\bibitem[{Uzunoglu and {\c{S}}ahin(2023)}]{uzunoglu-sahin-2023-benchmarking}
Arda Uzunoglu and G{\"o}zde {\c{S}}ahin. 2023.
\newblock \href {https://aclanthology.org/2023.ijcnlp-main.52} {Benchmarking procedural language understanding for low-resource languages: A case study on {T}urkish}.
\newblock In \emph{Proceedings of the 13th International Joint Conference on Natural Language Processing and the 3rd Conference of the Asia-Pacific Chapter of the Association for Computational Linguistics (Volume 1: Long Papers)}, pages 804--819, Nusa Dua, Bali. Association for Computational Linguistics.

\bibitem[{Wang et~al.(2023)Wang, Li, Yin, Wu, and Jia}]{wang2023emotional}
Xuena Wang, Xueting Li, Zi~Yin, Yue Wu, and Liu Jia. 2023.
\newblock \href {http://arxiv.org/abs/2307.09042} {Emotional intelligence of large language models}.

\bibitem[{Wei et~al.(2022{\natexlab{a}})Wei, Bosma, Zhao, Guu, Yu, Lester, Du, Dai, and Le}]{wei2022finetuned}
Jason Wei, Maarten Bosma, Vincent Zhao, Kelvin Guu, Adams~Wei Yu, Brian Lester, Nan Du, Andrew~M. Dai, and Quoc~V Le. 2022{\natexlab{a}}.
\newblock \href {https://openreview.net/forum?id=gEZrGCozdqR} {Finetuned language models are zero-shot learners}.
\newblock In \emph{International Conference on Learning Representations}.

\bibitem[{Wei et~al.(2022{\natexlab{b}})Wei, Tay, Bommasani, Raffel, Zoph, Borgeaud, Yogatama, Bosma, Zhou, Metzler, Chi, Hashimoto, Vinyals, Liang, Dean, and Fedus}]{wei2022emergent}
Jason Wei, Yi~Tay, Rishi Bommasani, Colin Raffel, Barret Zoph, Sebastian Borgeaud, Dani Yogatama, Maarten Bosma, Denny Zhou, Donald Metzler, Ed~H. Chi, Tatsunori Hashimoto, Oriol Vinyals, Percy Liang, Jeff Dean, and William Fedus. 2022{\natexlab{b}}.
\newblock \href {https://openreview.net/forum?id=yzkSU5zdwD} {Emergent abilities of large language models}.
\newblock \emph{Transactions on Machine Learning Research}.
\newblock Survey Certification.

\bibitem[{Xu et~al.(2023)Xu, Sun, Zheng, Geng, Zhao, Feng, Tao, and Jiang}]{xu2023wizardlm}
Can Xu, Qingfeng Sun, Kai Zheng, Xiubo Geng, Pu~Zhao, Jiazhan Feng, Chongyang Tao, and Daxin Jiang. 2023.
\newblock \href {http://arxiv.org/abs/2304.12244} {Wizardlm: Empowering large language models to follow complex instructions}.

\bibitem[{Zellers et~al.(2019)Zellers, Holtzman, Bisk, Farhadi, and Choi}]{zellers-etal-2019-hellaswag}
Rowan Zellers, Ari Holtzman, Yonatan Bisk, Ali Farhadi, and Yejin Choi. 2019.
\newblock \href {https://doi.org/10.18653/v1/P19-1472} {{H}ella{S}wag: Can a machine really finish your sentence?}
\newblock In \emph{Proceedings of the 57th Annual Meeting of the Association for Computational Linguistics}, pages 4791--4800, Florence, Italy. Association for Computational Linguistics.

\bibitem[{Zhan et~al.(2023)Zhan, Li, Wang, Luo, Feng, Kang, Hua, Qu, Soon, Sharma, Zukerman, Semnani-Azad, and Haffari}]{10.1145/3539618.3591877}
Haolan Zhan, Zhuang Li, Yufei Wang, Linhao Luo, Tao Feng, Xiaoxi Kang, Yuncheng Hua, Lizhen Qu, Lay-Ki Soon, Suraj Sharma, Ingrid Zukerman, Zhaleh Semnani-Azad, and Gholamreza Haffari. 2023.
\newblock \href {https://doi.org/10.1145/3539618.3591877} {Socialdial: A benchmark for socially-aware dialogue systems}.
\newblock In \emph{Proceedings of the 46th International ACM SIGIR Conference on Research and Development in Information Retrieval}, SIGIR '23, page 2712–2722, New York, NY, USA. Association for Computing Machinery.

\bibitem[{Zhang et~al.(2018)Zhang, Dinan, Urbanek, Szlam, Kiela, and Weston}]{zhang-etal-2018-personalizing}
Saizheng Zhang, Emily Dinan, Jack Urbanek, Arthur Szlam, Douwe Kiela, and Jason Weston. 2018.
\newblock \href {https://doi.org/10.18653/v1/P18-1205} {Personalizing dialogue agents: {I} have a dog, do you have pets too?}
\newblock In \emph{Proceedings of the 56th Annual Meeting of the Association for Computational Linguistics (Volume 1: Long Papers)}, pages 2204--2213, Melbourne, Australia. Association for Computational Linguistics.

\bibitem[{Zhang et~al.(2023)Zhang, Aljunied, Gao, Chia, and Bing}]{zhang2023m3exam}
Wenxuan Zhang, Sharifah~Mahani Aljunied, Chang Gao, Yew~Ken Chia, and Lidong Bing. 2023.
\newblock \href {http://arxiv.org/abs/2306.05179} {M3exam: A multilingual, multimodal, multilevel benchmark for examining large language models}.

\bibitem[{Zhang et~al.(2020)Zhang, Sun, Galley, Chen, Brockett, Gao, Gao, Liu, and Dolan}]{zhang-etal-2020-dialogpt}
Yizhe Zhang, Siqi Sun, Michel Galley, Yen-Chun Chen, Chris Brockett, Xiang Gao, Jianfeng Gao, Jingjing Liu, and Bill Dolan. 2020.
\newblock \href {https://doi.org/10.18653/v1/2020.acl-demos.30} {{DIALOGPT} : Large-scale generative pre-training for conversational response generation}.
\newblock In \emph{Proceedings of the 58th Annual Meeting of the Association for Computational Linguistics: System Demonstrations}, pages 270--278, Online. Association for Computational Linguistics.

\bibitem[{Zhao et~al.(2019)Zhao, Wang, Yatskar, Cotterell, Ordonez, and Chang}]{zhao-etal-2019-gender}
Jieyu Zhao, Tianlu Wang, Mark Yatskar, Ryan Cotterell, Vicente Ordonez, and Kai-Wei Chang. 2019.
\newblock \href {https://doi.org/10.18653/v1/N19-1064} {Gender bias in contextualized word embeddings}.
\newblock In \emph{Proceedings of the 2019 Conference of the North {A}merican Chapter of the Association for Computational Linguistics: Human Language Technologies, Volume 1 (Long and Short Papers)}, pages 629--634, Minneapolis, Minnesota. Association for Computational Linguistics.

\bibitem[{Zheng et~al.(2023)Zheng, Chiang, Sheng, Zhuang, Wu, Zhuang, Lin, Li, Li, Xing, Zhang, Gonzalez, and Stoica}]{zheng2023judging}
Lianmin Zheng, Wei-Lin Chiang, Ying Sheng, Siyuan Zhuang, Zhanghao Wu, Yonghao Zhuang, Zi~Lin, Zhuohan Li, Dacheng Li, Eric.~P Xing, Hao Zhang, Joseph~E. Gonzalez, and Ion Stoica. 2023.
\newblock \href {http://arxiv.org/abs/2306.05685} {Judging llm-as-a-judge with mt-bench and chatbot arena}.

\bibitem[{Zhou et~al.(2020)Zhou, Gao, Li, and Shum}]{10.1162/coli_a_00368}
Li~Zhou, Jianfeng Gao, Di~Li, and Heung-Yeung Shum. 2020.
\newblock \href {https://doi.org/10.1162/coli_a_00368} {{The Design and Implementation of XiaoIce, an Empathetic Social Chatbot}}.
\newblock \emph{Computational Linguistics}, 46(1):53--93.

\bibitem[{Zhou et~al.(2023)Zhou, Madaan, Potharaju, Gupta, McKee, Holtzman, Pujara, Ren, Mishra, Nematzadeh, Upadhyay, and Faruqui}]{zhou2023far}
Pei Zhou, Aman Madaan, Srividya~Pranavi Potharaju, Aditya Gupta, Kevin~R. McKee, Ari Holtzman, Jay Pujara, Xiang Ren, Swaroop Mishra, Aida Nematzadeh, Shyam Upadhyay, and Manaal Faruqui. 2023.
\newblock \href {http://arxiv.org/abs/2310.03051} {How far are large language models from agents with theory-of-mind?}

\end{thebibliography}

\clearpage
\begin{appendices}
\section{Data Preprocessing}
\label{appx:data_preprocessing}
\begin{table*}[t]
\centering
\begin{tabularx}{\textwidth}{llcX}
\toprule
\multicolumn{1}{c}{Task} & \multicolumn{1}{c}{Source} & Split & \multicolumn{1}{c}{Raw Categories} \\ \midrule
\multirow{15}{*}{Topic} & \multirow{4}{*}{K-SNS} & \multirow{4}{*}{Valid} & 일과 직업, 여가 생활, 시사/교육, 주거와 생활, 행사, 식음료, 개인 및 관계, 상거래(쇼핑), 미용과 건강 \\
 &  &  & work, leisure, news/education, living, event, food, relationship, shopping, health and beauty \\ \cmidrule{2-4}
 & \multirow{6}{*}{K-TDD} & \multirow{6}{*}{Valid} & 사회 이슈, 식음료, 가족, 교육, 건강, 계절/날씨, 타 국가 이슈, 교통, 방송/연예, 군대, 여행, 회사/아르바이트, 게임, 연애/결혼, 영화/만화, 스포츠/레저, 미용, 반려동물, 상거래 전반, 주거와 생활 \\ 
 &  &  & domestic issue, food, family, education, health, weather, international issue, transportation, entertainment, military, travel, job, game, love/marriage, movie/cartoon, sports, beauty, pet, shopping, living \\ \cmidrule{2-4}
 & \multirow{5}{*}{SD (K)} & \multirow{5}{*}{-} & 비리/부패, 여행, 회사 업무, 음식, 농업, 학교 생활, 범죄/테러, 상거래, 재난 피해자, 빈곤 구호, 아동 실종, 일상 생활 \\
 &  &  & police-corruption, tourism, office-affairs, food, farming, school-life, counter-terrorism/anti-crime, sale, disaster-victims, poverty-assistance, child-missing, life-trivial \\ \midrule
\multirow{12}{*}{Emotion} & \multirow{2}{*}{K-ED} & \multirow{2}{*}{Valid} & 불안, 슬픔, 당황, 분노, 상처, 기쁨 \\
 &  &  & anxiety, sadness, embarrassment, anger, hurt, happiness \\ \cmidrule{2-4}
 & \multirow{2}{*}{DD (K)} & \multirow{2}{*}{Test} & 감정 없음, 화남, 혐오, 두려움, 기쁨, 슬픔, 놀람 \\
 &  &  & no emotion, anger, disgust, fear, happiness, sadness, surprise \\ \cmidrule{2-4}
 & \multirow{8}{*}{ED (K)} & \multirow{8}{*}{Test} & 신남, 화남, 자랑스러움, 슬픔, 짜증, 감사, 외로움, 두려움, 무서움, 죄책감, 감명, 혐오, 희망, 자신감, 분노, 걱정, 기대, 기쁨, 향수, 실망, 준비, 질투, 만족, 충격, 당황, 배려, 신뢰, 수치, 걱정, 믿음 \\
 &  &  & excited, angry, proud, sad, annoyed, grateful, lonely, afraid, terrified, guilty, impressed, disgusted, hopeful, confident, furious, anxious, anticipating, joyful, nostalgic, disappointed, prepared, jealous, content, devastated, embarrassed, caring, trusting, ashamed, apprehensive, faithful \\ \midrule
\multirow{6}{*}{Relation} & \multirow{2}{*}{SD$_\text{Dst}$ (K)} & \multirow{2}{*}{-} & 이웃, 낯선 사람, 친구, 직장, 연인, 가족 \\
 &  &  & neighborhood, stranger, friend, working, romantic, family \\ \cmidrule{2-4}
 & \multirow{4}{*}{SD$_\text{Rel}$ (K)} & \multirow{4}{*}{-} & 동료-동료, 상사-부하, 멘토-멘티, 지휘관-병사, 연인 또는 부부, 고객-직원, 손윗사람-손아랫사람, 학생-선생님 \\
 &  &  & peer-peer, chief-subordinate, mentor-mentee, commander-soldier, partner-partner, customer-server, elder-junior, student-professor \\ \midrule
\multirow{4}{*}{Location} & \multirow{4}{*}{SD (K)} & \multirow{4}{*}{-} & 학교, 가게, 집, 호텔, 공공장소, 식당, 전화통화, 경찰서, 사무실, 난민 캠프 \\
 &  &  & school, store, home, hotel, open-area, restaurant, online, police-station, office, refugee-camp \\ \midrule
\multirow{4}{*}{Dialog Act} & \multirow{2}{*}{K-TDD} & \multirow{2}{*}{Valid} & 지시, 단언, 언약, 표현 \\
 &  &  & directive, assertive, commissive, expressive \\ \cmidrule{2-4}
 & \multirow{2}{*}{DD (K)} & \multirow{2}{*}{Test} & 알림, 질문, 지시, 언약 \\
 &  &  & inform, question, directive, commissive \\ \bottomrule
\end{tabularx}%
\caption{Detailed description of dialogue comprehension task suite. We enumerate the original categories of raw corpora in both Korean and English. We use the class names in Korean to all of our tasks.}
\label{tab:detailed_desc}
\end{table*}

We elucidate the detailed statistics and preprocessing procedures of raw corpora in our dialogue comprehension task suite.
For native Korean corpora, we modify the class names to make them represent the dialogue contents precisely.
For translated corpora, we also translate the class names to Korean.
See Table~\ref{tab:detailed_desc} for more details.

\subsection{Topic Classification}
\paragraph{Korean SNS}
The original corpus contains 200k dialogues, each annotated with one of 9 topic categories.
We remove the 주거와 생활(\textit{living}), 행사 (\textit{event}), and 개인 및 관계 (\textit{relationship}) categories due to their ambiguity in distinction from other categories.
From these, we then randomly select 200 dialogue examples for each class.

\paragraph{Korean Thematic Daily Dialogues}
The raw corpus comprises 10,962 dialogues, annotated across 20 topic categories
We exclude the 가족(\textit{family}) class due to semantic overlap with other categories, resulting in a refined list of 19 classes.
From each class, we randomly sample 100 dialogue examples.

\paragraph{SocialDial (Korean): Topic}
From the initial 12 topic classes, we eliminate 7 categories with fewer than 50 examples each: \textit{police-corruption}, \textit{tourism}, \textit{farming}, \textit{counter-terrorism/anti-crime}, \textit{disaster-victims}, \textit{poverty-assistance}, and \textit{child-missing}.
Additionally, we exclude the \textit{life-trivial} class due to its semantic overlap with other categories, ultimately yielding 4 classes.
For each of these classes, we randomly sample 100 dialogue examples.

\subsection{Emotion Recognition}
\paragraph{Korean Emotional Dialogues}
The raw corpus encompasses 6,641 dialogues, each with up to three turns (i.e., six utterances), and is annotated with 6 emotion categories.
We retain these classes without modification.
Furthermore, we randomly sample 200 three-turn dialogues from each class.

\paragraph{DailyDialog (Korean): Emotion}
The raw corpus comprise 7 emotion categories.
We exclude the \textit{disgust} and \textit{fear} categories, which have fewer than 50 examples each, resulting in 5 classes.
From each class, we randomly sample 94 dialogue examples.

\paragraph{Empathetic Dialogues (Korean): Emotion}
The raw corpus encompasses 32 emotion categories, which we consolidate into \textit{positive} and \textit{negative} classes, excluding \textit{surprise} and \textit{sentimental} categories.
The 긍정(\textit{positive}) class amalgamates 14 categories: \textit{excited}, \textit{proud}, \textit{grateful}, \textit{impressed}, \textit{hopeful}, \textit{confident}, \textit{anticipating}, \textit{joyful}, \textit{nostalgic}, \textit{prepared}, \textit{content}, \textit{caring}, \textit{trusting}, and \textit{faithful}.
Conversely, the 부정(\textit{negative}) class comprises 16 categories: \textit{angry}, \textit{sad}, \textit{lonely}, \textit{afraid}, \textit{terrified}, \textit{guilty}, \textit{disgusted}, \textit{furious}, \textit{anxious}, \textit{disappointed}, \textit{jealous}, \textit{devastated}, \textit{embarrassed}, \textit{ashamed}, and \textit{apprehensive}.
We randomly sample 1,000 dialogue examples from each polarity.

\subsection{Relation Classification}
\paragraph{SocialDial (Korean): Social Distance}
The raw corpus is grouped into 6 social distance categories.
We omit the \textit{neighborhood} and \textit{romantic} categories due to their having fewer than 50 examples, thereby utilizing 4 classes.
From each class, we randomly sample 131 dialogue examples.

\paragraph{SocialDial (Korean): Social Relation}
The raw corpus encompasses 8 distinct social relation categories.
Initially, we amalgamate the \textit{commander-soldier} category into \textit{chief-subordinate} and the \textit{mentor-mentee} into \textit{student-professor}, respectively.
Subsequently, the \textit{partner-partner} category is omitted due to comprising fewer than 50 examples.
Further, we exclude the \textit{peer-peer} and \textit{elder-junior} categories due to inconsistencies in the translation of formality forms.
Consequently, this refinement results in 3 classes, from each of which we randomly sample 110 dialogue examples.

\subsection{Location Classification}
\paragraph{SocialDial (Korean): Location}
The original corpus encompasses 10 location categories.
We exclude the \textit{home} and \textit{open-area} categories due to their indistinct boundaries with other categories and omit the \textit{hotel}, \textit{online}, \textit{police-station}, and \textit{refugee-camp} categories, each containing fewer than 50 examples.
This refinement results in 4 classes.
From each class, we randomly sample 94 dialogue examples.

\subsection{Dialog Act Classification}
\paragraph{Korean Thematic Daily Dialogues}
The raw corpus consists of 4 dialog act classes.
We employ these classes without modification and randomly sample 130 dialogue examples from each class.

\paragraph{DailyDialog (Korean): Act}
The raw corpus is composed of 4 dialog act classes.
We randomly sample 250 dialogue examples from each class.

\section{Prompts}
\label{appx:prompts}
We illustrate prompt examples used in our experiments along with the line-by-line translations.

\subsection{Direct Prompting}
\begin{tcolorbox}[colback=green!5,colframe=green!40!black,title=Direct Prompting,enhanced,breakable]
화자2: 그동안 많이 힘들었겠군요.\\
Speaker 2: I'm sorry you've been through so much.\\
화자1: 맞아. 근데 이젠 가족들에게 속마음을 털어놓을 수 있어 기뻐.\\
Speaker 1: Yes, but I'm glad I can open up to my family now.\\
화자2: 가족들에게 마음을 털어 놓아 편안하시군요.\\
Speaker 2: It's good to hear that you feel comfortable opening up to your family.\\
\\
질문: 대화에서 화자1이 느끼는 감정은 무엇인가?\\
Question: What is Speaker 1 feeling in this conversation?\\
정답:\\
Answer:
\end{tcolorbox}

\subsection{Direct Prompting with Class Descriptions}
\begin{tcolorbox}[colback=green!5,colframe=green!40!black,title=Direct Prompting with Class Descriptions,enhanced,breakable]
$\left[\text{대화}\right]$\\
$\left[\text{Dialogue}\right]$\\
화자2: 잠깐 일시적으로 추워진 거래 키키\\
Speaker 2: It has just become temporarily cold they say, lol.\\
화자1: 아 진짜? 다행이다\\
Speaker 1: Oh, really? Thank God.\\
화자1: 한파라고 해서 진짜 오잉 했잖아!\\
Speaker 1: I was so freaking out when they said it was cold surge!\\
화자2: 나도 한파 주의보 문자 와서 당황함\\
Speaker 2: I was also confused when I got the cold surge warning.\\
\\
$\left[\text{보기}\right]$\\
$\left[\text{Choices}\right]$\\
지시: 상대에게 충고, 제안, 명령, 요구, 질문, 부탁 등을 하는 발화\\
Directive: an utterance that gives advice, suggestions, orders, demands, questions, favors, etc. to the other.\\
단언: 자신의 의견을 진술, 주장하거나 상대의 의견을 반박하는 발화\\
Assertive: an utterance that states or asserts one's opinion or refutes the other's opinion.\\
언약: 상대와 약속을 하거나 상대의 요청을 거절하는 발화\\
Commissive: an utterance in which one make a promise to or refuse a request from the other.\\
표현: 인사, 감사, 사과, 긍정 및 부정 감정 표현 등을 하는 발화\\
Expressive: an utterance that gives greetings, thanks, apologies, expressions of positive and negative emotions, etc.\\
\\
질문: 보기 중 대화의 마지막 발화의 의도로 가장 알맞은 것은?\\
Question: Which of the choices best describes the intent of the last utterance?\\
정답:\\
Answer:
\end{tcolorbox}

\subsection{Option Prompting}
\begin{tcolorbox}[colback=green!5,colframe=green!40!black,title=Option Prompting,enhanced,breakable]
$\left[\text{대화}\right]$\\
$\left[\text{Dialogue}\right]$\\
화자1: 이렇게 늦게까지 뭐하세요?\\
Speaker 1: What are you doing up so late?\\
화자2: 요리하고 있었어요! 당신은요?\\
Speaker 2: I've been cooking! You?\\
화자1: 개 산책시키고 있어요.\\
Speaker 1: I'm walking the dog.\\
화자2: 이렇게 늦게까지! 그냥 요리 연습 중이에요. 몇 살이에요?\\
Speaker 2: This late! I'm just practicing cooking. How old are you?\\
화자1: 이 늦은 시간에 타코를 만드시네요.\\
Speaker 1: You're making tacos at this late hour.\\
화자2: 네! 나 타코 좋아해요! 제가 싫어하는 건 별로 없어요, 23살이에요.\\
Speaker 2: Yeah! I love tacos! There's not much I don't like, I'm 23 years old.\\
화자1: 어디 사세요?\\
Speaker 1: Where do you live?\\
화자2: 지금은 오리건주에 살고 있지만 올해 전 세계를 돌아다녔어요.\\
Speaker 2: I live in Oregon right now, but I've been traveling around the world this year.\\
화자1: 앨라배마에 가보셨어요?\\
Speaker 1: Have you been to Alabama?\\
화자2: 네. 거기서 먹은 타코가 정말 맛있었어요! 어디에 사세요?\\
Speaker 2: Yeah. I loved the tacos there! Where do you live?\\
화자1: 몽고메리에 살고 있어요.\\
Speaker 1: I live in Montgomery.\\
화자2: 요트를 거기 근처에 보관하고 있어요. 지금은 빌려주고 있어요.\\
Speaker 2: I keep my yacht near there, and I'm renting it out now.\\
화자1: 요트가 멋지네요.\\
Speaker 1: That's a nice yacht.\\
\\
$\left[\text{보기}\right]$\\
$\left[\text{Choices}\right]$\\
1) 3개월 후에 세 쌍둥이를 출산할 예정입니다.\\
1) I am expecting triplets in three months.\\
2) 빨간색에 파란색 줄무늬가 있어 레이스할 때 반짝반짝 빛납니다.\\
2) It's red with blue stripes, so it sparkles when I race.\\
3) 저는 여행을 정말 좋아합니다.\\
3) I really like traveling.\\
4) 가는 곳마다 온갖 종류의 음식을 다 먹어봤어요.\\
4) I've tried every kind of food everywhere I've been.\\
\\
질문: 보기 중 화자2에 관한 서술로 옳은 것은?\\
Question: Which statement about speaker 2 in the choices is correct?\\
정답:\\
Answer:
\end{tcolorbox}

\subsection{Response Selection Prompting}
\begin{tcolorbox}[colback=green!5,colframe=green!40!black,title=Response Selection Prompting,enhanced,breakable]
화자2: 요새 샐러드 파는 가게\\
Speaker 2: Salad shops these days\\
화자2: 많아졌따\\
Speaker 2: There's a lot of them.\\
화자1: 맞아 그리고 진짜\\
Speaker 1: Yeah, and really\\
화자1: 퀄리티도 좋더라 요새는\\
Speaker 1: The quality is good, too.\\
화자2: 응 맨날 사먹고 싶게 생겼어\\
Speaker 2: Yeah, it looks so good that I'd like to eat it everyday.\\
화자2: 샐러드인데도\\
Speaker 2: Even though it's a salad\\
화자1: 근데 가격도\\
Speaker 1: But the price\\
화자1: 비싸더라고...\\
Speaker 1: It is expensive...\\
화자2:\\
Speaker 2:
\end{tcolorbox}

\section{Detailed Results}
\label{appx:detailed_results}
\subsection{Results on Individual Tasks in Dialogue Comprehension}
We provide the detailed results for individual tasks of dialogue comprehension in Table~\ref{tab:result-comprehension-all}.
Each task consists of examples from the same data sources.
The acronyms of datasets not defined in Table~\ref{tab:result-response-selection} are defined as follows:

\begin{itemize}
    \item{SD$_\text{Dst}$ (K): social distance classes from SD (K)}
    \item{SD$_\text{Rel}$ (K): social relation classes from SD (K)}
    \item{K-DS: Korean Dialogue Summary}
\end{itemize}

\subsection{Reliability of Human Evaluation}

\begin{table}[ht]
\centering
\begin{tabular}{@{}llll@{}}
\toprule
\multicolumn{3}{c}{Task} & \multicolumn{1}{c}{$\kappa$} \\ \midrule
\multirow{14}{*}{\rotatebox[origin=c]{90}{\small{Dialogue Comprehension}}} & \multirow{3}{*}{Topic} & K-SNS & 0.800 \\
 &  & K-TDD & 0.865 \\
 &  & SD (K) & 0.946 \\ \cmidrule(l){2-4} 
 & Location & SD (K) & 0.794 \\ \cmidrule(l){2-4} 
 & \multirow{2}{*}{Relation} & SD$_\text{Dst}$ (K) & 0.540 \\
 &  & SD$_\text{Rel}$ (K) & 0.900 \\ \cmidrule(l){2-4} 
 & \multirow{3}{*}{Emotion} & K-ED & 0.628 \\
 &  & DD (K) & 0.536 \\
 &  & ED (K) & 0.678 \\ \cmidrule(l){2-4} 
 & \multirow{2}{*}{Dialog Act} & K-TDD & 0.415 \\
 &  & DD (K) & 0.370 \\ \cmidrule(l){2-4} 
 & \multirow{3}{*}{Fact} & K-DS & 0.982 \\
 &  & PC (K) & 0.706 \\
 &  & ED (K) & 1.000 \\ \midrule
\multirow{7}{*}{\rotatebox[origin=c]{90}{\small{Response Selection}}} &  & K-SNS & 0.815 \\
 &  & K-TDD & 0.965 \\
 &  & K-ED & 0.983 \\
 &  & PC (K) & 0.829 \\
 &  & DD (K) & 0.838 \\
 &  & ED (K) & 0.830 \\
 &  & SD (K) & 0.838 \\ \bottomrule
\end{tabular}
\caption{Inter-rater agreements of human evaluators for each task.}
\label{tab:fleiss_kappa}
\end{table}

We calculate Fleiss' kappa to estimate the inter-rater agreements of the three human evaluators (Table~\ref{tab:fleiss_kappa}).
We observe almost perfect ($>0.8$) and substantial ($>0.6$) agreements in most tasks, whereas we observe moderate ($>0.4$) agreements in Relation: SD$_\text{Dst}$ (K), Emotion: DD (K) and Dialog Act: K-TDD, and fair ($>0.2$) agreement in Dialog Act: DD (K).
We speculate the subjective nature of emotion led to the low performance and agreements of human evaluators in emotion recognition tasks.
For dialog act classification, it is because the verbalized class names are defined academically and connote several hyponyms, which is hard for humans to precisely understand without linguistic knowledge.

\begin{landscape}
\begin{table}[p]
\centering
\resizebox{0.9\linewidth}{!}{%
\begin{tabular}{@{}l|ccc|c|cc|ccc|cc|ccc|c@{}}
\toprule
\multicolumn{1}{c|}{\multirow{2}{*}{Model}} & \multicolumn{3}{c|}{Topic} & Location & \multicolumn{2}{c|}{Relation} & \multicolumn{3}{c|}{Emotion} & \multicolumn{2}{c|}{Dialog Act} & \multicolumn{3}{c|}{Fact} & \multirow{2}{*}{Average} \\ \cmidrule(lr){2-15}
\multicolumn{1}{c|}{} & K-SNS & K-TDD & SD (K) & SD (K) & SD$_\text{Dst}$ (K) & SD$_\text{Rel}$ (K) & K-ED & DD (K) & ED (K) & K-TDD & DD (K) & K-DS & PC (K) & ED (K) &  \\ \midrule
Random & 16.7 & 5.3 & 25.0 & 25.0 & 25.0 & 33.3 & 16.7 & 20.0 & 50.0 & 25.0 & 25.0 & 25.0 & 25.0 & 25.0 & 24.4 \\ \midrule
XGLM 564M & 30.8 & 28.1 & 33.5 & 52.1 & 30.9 & 40.6 & 32.5 & 28.1 & 50.6 & 25.0 & 24.4 & 25.6 & 24.5 & 25.4 & 32.3 \\
XGLM 1.7B & 30.3 & 28.4 & 31.8 & 48.1 & 32.6 & 34.2 & 43.8 & 24.5 & 50.1 & 24.6 & 25.4 & 25.6 & 24.6 & 25.5 & 32.1 \\
XGLM 2.9B & 37.8 & 32.5 & 41.5 & 45.7 & 26.9 & 55.2 & 49.1 & 33.4 & 50.5 & 24.8 & 25.2 & 25.6 & 24.6 & 25.5 & 35.6 \\
XGLM 4.5B & 28.8 & 29.5 & 38.8 & 61.4 & 26.5 & 45.5 & 48.6 & 31.9 & 50.1 & 25.2 & 26.4 & 25.6 & 24.6 & 25.5 & 34.9 \\
XGLM 7.5B & 35.4 & 33.9 & 47.3 & 69.4 & 26.9 & 57.6 & 51.4 & 37.0 & 63.2 & 24.8 & 25.0 & 25.6 & 23.8 & 25.5 & 39.1 \\
LLaMA 7B & 24.8 & 26.9 & 27.0 & 35.9 & 33.0 & 54.8 & 44.0 & 27.0 & 67.2 & 22.9 & 24.5 & 25.6 & 25.7 & 27.0 & 33.3 \\
LLaMA 13B & 25.1 & 28.7 & 34.0 & 42.3 & 38.0 & 37.0 & 39.7 & 23.0 & 50.4 & 25.0 & 22.1 & 32.6 & 31.6 & 38.4 & 33.4 \\
WizardLM 7B & 16.7 & 5.3 & 25.0 & 25.0 & 25.0 & 33.3 & 16.7 & 20.0 & 50.0 & 25.0 & 25.0 & 23.9 & 23.2 & 25.8 & 24.3 \\
WizardLM 13B & 28.5 & 29.2 & 28.0 & 42.6 & 38.9 & 33.9 & 31.2 & 27.0 & 50.0 & 24.0 & 25.1 & 36.6 & 34.9 & 32.1 & 33.0 \\
LLaMA-2 7B & 32.3 & 33.6 & 46.5 & 75.8 & 38.9 & 74.2 & 49.6 & 38.1 & 51.4 & 24.0 & 25.4 & 30.9 & 33.3 & 32.1 & 41.9 \\
LLaMA-2 13B & 32.9 & 37.2 & 37.8 & 78.2 & 47.7 & 58.8 & 54.4 & 35.5 & 73.9 & 24.8 & 24.8 & 26.7 & 32.5 & 54.9 & 44.3 \\
LLaMA-2-Chat 7B & 29.9 & 24.9 & 38.3 & 67.3 & 40.1 & 48.5 & 36.0 & 28.5 & 65.5 & 26.7 & 27.4 & 39.0 & 27.3 & 31.9 & 38.0 \\
LLaMA-2-Chat 13B & 34.9 & 29.9 & 46.3 & 74.7 & 53.6 & 38.8 & 47.0 & 23.6 & 83.4 & 22.1 & 25.0 & 39.8 & 27.8 & 56.9 & 43.1 \\
Falcon 7B & 16.9 & 16.7 & 25.5 & 29.5 & 26.9 & 43.3 & 28.6 & 20.0 & 50.1 & 24.8 & 25.1 & 25.6 & 24.6 & 25.5 & 27.4 \\
Falcon-Inst 7B & 19.1 & 21.4 & 27.0 & 26.6 & 26.2 & 48.2 & 38.6 & 20.4 & 50.0 & 24.4 & 24.7 & 25.5 & 24.6 & 25.4 & 28.7 \\
Mistral 7B & 26.8 & 36.2 & 39.5 & 76.9 & 34.9 & 58.5 & 54.3 & 38.3 & 83.7 & 24.8 & 25.1 & 79.0 & 45.8 & 80.2 & 50.3 \\
Mistral-Inst 7B & 26.3 & 31.3 & 26.0 & 39.9 & 42.6 & 63.3 & 47.7 & 23.4 & 57.4 & 26.2 & 27.6 & 46.9 & 42.2 & 69.9 & 40.8 \\
CodeLLaMA 7B & 24.8 & 28.6 & 38.8 & 52.7 & 25.6 & 53.9 & 49.2 & 37.9 & 53.2 & 21.9 & 25.6 & 42.8 & 40.5 & 42.1 & 38.4 \\
CodeLLaMA 13B & 34.5 & 33.4 & 33.5 & 63.3 & 31.7 & 76.4 & 57.0 & 49.2 & 83.3 & 25.0 & 26.0 & 36.8 & 28.0 & 30.6 & 43.5 \\
CodeLLaMA-Inst 7B & 27.1 & 28.3 & 41.0 & 55.3 & 26.7 & 56.4 & 51.4 & 37.9 & 67.4 & 25.2 & 28.4 & 46.2 & 35.6 & 59.7 & 41.9 \\
CodeLLaMA-Inst 13B & 34.3 & 33.3 & 31.0 & 63.3 & 35.3 & 79.7 & 56.8 & 51.3 & 83.6 & 25.0 & 26.9 & 46.0 & 42.1 & 60.0 & 47.7 \\
Qwen 7B & 33.3 & 35.7 & 47.3 & 64.1 & 25.6 & 33.3 & 31.8 & 34.7 & 81.9 & 25.0 & 25.0 & 43.7 & 37.2 & 54.3 & 40.9 \\
Qwen 14B & 42.3 & 45.4 & 48.3 & 73.7 & 28.4 & 55.5 & 50.8 & 43.0 & 85.5 & 25.0 & 30.4 & 92.3 & 48.2 & 91.0 & 54.3 \\
Qwen-Chat 7B & 34.6 & 27.4 & 46.8 & 48.4 & 25.6 & 51.5 & 25.7 & 33.0 & 53.0 & 26.0 & 24.9 & 63.8 & 44.2 & 66.1 & 40.8 \\
Qwen-Chat 14B & 36.8 & 39.5 & 49.0 & 68.6 & 28.4 & 54.6 & 44.3 & 33.0 & 87.3 & 30.0 & 27.3 & 99.0 & 54.3 & 95.3 & 53.4 \\
Polyglot-Ko 1.3B & 32.6 & 29.7 & 33.0 & 61.7 & 31.7 & 46.4 & 52.0 & 32.3 & 50.0 & 24.4 & 25.1 & 25.7 & 24.5 & 25.7 & 35.3 \\
Polyglot-Ko 3.8B & 37.5 & 35.2 & 36.0 & 58.8 & 26.9 & 63.9 & 57.6 & 36.6 & 50.0 & 25.2 & 25.3 & 25.5 & 24.8 & 25.6 & 37.8 \\
Polyglot-Ko 5.8B & 19.5 & 26.6 & 43.0 & 59.3 & 32.6 & 47.3 & 51.8 & 37.4 & 50.2 & 27.7 & 25.0 & 25.8 & 24.6 & 25.5 & 35.4 \\
Polyglot-Ko 12.8B & 33.0 & 35.3 & 41.8 & 61.7 & 36.5 & 57.9 & 54.0 & 41.7 & 65.9 & 25.2 & 24.7 & 23.8 & 23.2 & 26.0 & 39.3 \\
KoAlpaca 5.8B & 31.9 & 20.9 & 46.5 & 47.1 & 29.8 & 33.3 & 39.9 & 29.6 & 51.5 & 25.0 & 21.6 & 24.2 & 24.2 & 24.3 & 32.1 \\
KoAlpaca 12.8B & 36.2 & 33.9 & 56.3 & 70.5 & 37.2 & 51.2 & 55.1 & 43.0 & 82.1 & 25.0 & 23.6 & 22.0 & 23.6 & 25.4 & 41.8 \\
KORani-v1 13B & 25.8 & 36.0 & 38.3 & 73.1 & 42.4 & 49.1 & 49.7 & 34.7 & 71.5 & 21.0 & 27.0 & 23.0 & 24.6 & 25.2 & 38.7 \\
KORani-v2 13B & 21.4 & 37.0 & 33.0 & 68.1 & 29.6 & 61.8 & 37.0 & 26.4 & 53.7 & 25.6 & 25.0 & 35.1 & 33.8 & 34.8 & 37.3 \\
KORani-v3 13B & 24.9 & 40.7 & 38.3 & 69.7 & 31.5 & 51.5 & 38.4 & 26.2 & 80.2 & 26.9 & 26.5 & 32.2 & 28.9 & 47.4 & 40.2 \\ \midrule
Human & 72.0 & 82.7 & 96.0 & 86.0 & 58.7 & 88.0 & 66.0 & 55.3 & 80.0 & 51.3 & 58.0 & 99.3 & 80.7 & 100.0 & 76.7 \\ \bottomrule
\end{tabular}%
}
\caption{Results for individual tasks in dialogue comprehension.}
\label{tab:result-comprehension-all}
\end{table}
\end{landscape}

\end{appendices}

\end{document}